\documentclass[journal,11pt,draftcls,onecolumn]{IEEEtran}

\usepackage{hero}

\usepackage{color}
\usepackage{amssymb,amstext,amsmath,amsfonts}
\usepackage{graphicx}
\usepackage{turnstile}
\usepackage{subfigure}
\usepackage[ruled]{algorithm2e}
\usepackage{enumerate}
\usepackage{tikz}
\tikzstyle{matched}=[circle, draw, fill=black!30,
inner sep=0pt, minimum width=4pt,fill=black!100]

\usepackage[margin=1in]{geometry}

\renewcommand{\r}{\mbox{\rm r}}
\def\ra{\mbox{\rm a}}
\newcommand{\qed}{\mbox{} \hfill $\Box$ }

\def\r{{\rm r}}
\def\ra{\mbox{\rm a}}
\def\mN{{\mathbb N}}
\def\mX{{\mathbb X}}
\def\mU{{\mathbb U}}
\def\mY{{\mathbb Y}}
\def\mZ{{\mathbb Z}}

\def\Var{\text{Var}}

\newtheorem{assumption}{Assumption}

\begin{document}

\title{Two-stage Sampling, Prediction and Adaptive Regression via Correlation Screening (SPARCS)}

\author{Hamed~Firouzi,~\IEEEmembership{Member,~IEEE,}
        Alfred~O.~Hero~III,~\IEEEmembership{Fellow,~IEEE,}
        Bala~Rajaratnam,~\IEEEmembership{Member,~IEEE}
\thanks{Parts of this work were presented at the 2013 Conference on Artificial Intelligence and Statistics  (AISTATS) and at the 2013 IEEE International Workshop on Computational Advances in
Multi-Sensor Adaptive Processing (CAMSAP).}
\thanks{This research was partially supported by the US National Science Foundation under grants CCF-1217880, DMS-CMG-1025465, AGS-1003823, DMS-1106642, and DMS-CAREER-1352656, by the US Air Force Office of Scientific Research under grant FA9550-13-1-0043, and by the US Army Research Office under grant W911NF-15-1-0479.}}
\IEEEpubid{0000--0000/00\$00.00 \copyright\ 2014 IEEE}

\maketitle

\begin{abstract}
This paper proposes a general adaptive procedure for budget-limited predictor design in high dimensions called two-stage Sampling, Prediction and Adaptive Regression via Correlation Screening (SPARCS). SPARCS can be applied to high dimensional prediction problems in experimental science, medicine, finance, and engineering, as illustrated by the following. Suppose one wishes to run a sequence of experiments to learn a sparse multivariate predictor of a dependent variable $Y$ (disease prognosis for instance) based on a $p$ dimensional set of independent variables $\bX=[X_1,\ldots, X_p]^T$ (assayed biomarkers). Assume that the cost of acquiring the full set of  variables $\bX$ increases linearly in its dimension. SPARCS  breaks the data collection into two stages in order to achieve an optimal tradeoff between sampling cost and predictor performance. In the first stage we collect a few ($n$) expensive samples $\{y_i,\bx_i\}_{i=1}^n$, at the full dimension $p\gg n$ of $\bX$, winnowing the number of variables down to a smaller dimension $l < p$ using a type of cross-correlation or regression coefficient screening. In the second stage we collect a larger number $(t-n)$ of cheaper samples of the $l$ variables that passed the screening of the first stage. At the second stage, a low dimensional predictor is constructed by solving the standard regression problem using all $t$ samples of the selected variables.
SPARCS is an adaptive online algorithm that implements false positive control on the selected variables, is well suited to small sample sizes, and is scalable to high dimensions. We establish asymptotic bounds for the Familywise Error Rate (FWER), specify high dimensional convergence rates for support recovery, and establish optimal sample allocation rules to the first and second stages. 
\end{abstract}

\begin{IEEEkeywords}
high dimensional regression, predictive modeling, model selection, thresholding, two-stage prediction, graphical models.
\end{IEEEkeywords}


\IEEEpeerreviewmaketitle


\section{Introduction} \label{sec:intro}

Much effort has been invested in the sparse regression problem where the objective is to  learn a sparse linear predictor from training data $\{ y_i, x_{i1},x_{i2},\ldots,x_{ip} \}_{i=1}^n$ where the number $p$ of predictor variables is much larger that the number $n$ of training samples. Applications in science and engineering where such ``small $n$ large $p$" problems arise include: sparse signal reconstruction \cite{Candes&etal:05,donoho2006compressed}; channel estimation in multiple antenna wireless communications  \cite{Hassibi&Hochwald:2003,biguesh2006training}; text processing of internet documents  \cite{forman2003extensive,ding2002adaptive}; gene expression array analysis \cite{golub1999molecular}; combinatorial chemistry \cite{suh2009visualization}; environmental sciences \cite{rong2011practical}; and others  \cite{guyon2003introduction}. In this $n \ll p$ setting training a linear predictor becomes difficult due to rank deficient normal equations, overfitting errors, and high computational complexity.
\IEEEpubidadjcol

A large number of methods for solving the sparse regression problem have been proposed. These include methods that simultaneously perform variable selection, and predictor design, and the methods that perform these two operations separately. The former class of methods includes, for example,  least absolute shrinkage and selection operator (LASSO), elastic LASSO, and group LASSO \cite{guyon2003introduction,tibshirani1996regression,efron2004least,buehlmann2006boosting,
yuan2005model,friedman2001elements, buhlmann2011statistics}. The latter class of methods includes sequential thresholding approaches such as sure independence screening (SIS); and marginal regression \cite{fan2008sure,genovese2009revisiting,genovese2012comparison,fan2010sure}. All of these methods are offline in the sense that they learn the predictor from a batch of precollected samples of all the variables.   In this paper we propose an online framework, called two-stage Sampling, Prediction and Adaptive Regression via Correlation Screening (SPARCS), which unequally and adaptively samples the variables in the process of constructing the predictor. One of the principal results of this paper is that, as compared under common sampling budget constraints, the proposed SPARCS method results in better prediction performance than offline methods.

Specifically, the SPARCS method for online sparse regression operates in two-stages. 
The first stage, which we refer to as the \emph{SPARCS screening stage}, collects a small number of full dimensional samples and performs variable selection on them. Variable selection at the SPARCS screening stage can be performed in one of two ways, i.e., by screening the sample cross-correlation between $Y$ and $\bX$, as in sure independence screening (SIS), or by thresholding the generalized Ordinary Least Squares (OLS) solution
, which we propose in this paper and we refer to as predictive correlation screening (PCS).  The second stage of SPARCS, referred to as the \emph{SPARCS regression stage}, collects a larger number of reduced dimensional samples, consisting only of the variables selected at the first stage, and regresses the responses on the the selected variables to build the predictor.

We establish the following theoretical results on SPARCS. First, under a sparse correlation assumption, we establish a Poisson-like limit theorem for the number of variables that pass the SPARCS screening stage as $p\rightarrow\infty$ for fixed $n$. This yields a Poisson approximation to the probability of false discoveries that is accurate for small $n$ and very large $p$. The Poisson-like limit theorem also specifies a phase transition threshold for the false discovery probability.
Second, with $n$, the number of samples in the first stage, and $t$,  the total number of samples, we establish that $n$ needs only be of order $\log p$ for SPARCS to succeed in recovering the support set of the optimal OLS predictor. Third, given a cost-per-sample that is linear in the number of assayed variables, we show that the optimal value of  $n$ is on the order of $\log t$. The above three results, established for our SPARCS framework, can be compared to theory for correlation screening \cite{hero2011large,hero2012hub}, support recovery for multivariate LASSO \cite{obozinski2011support}, and optimal exploration vs. exploitation allocation in multi-armed bandits \cite{audibert2007tuning}.

SPARCS can of course also be applied offline. When implemented in this way, it can be viewed as an alternative to LASSO-type regression methods \cite{tibshirani1996regression,paul2008preconditioning,wainwright2009sharp,huang2011variable,wauthier2013comparative}.
LASSO based methods try to perform simultaneous variable selection and regression via minimizing an $\ell_1$-regularized Mean Squared Error (MSE) objective function. Since the $\ell_1$-regularized objective function is not differentiable, such an optimization is computationally costly, specially for large $p$. Several approaches such as LARS \cite{efron2004least,khan2007robust,hesterberg2008least}, gradient projection
methods \cite{figueiredo2007gradient,quattoni2009efficient}, interior point methods \cite{kim2007interior,koh2007interior} and active-set-type algorithms \cite{kim2010fast,wen2010fast,wen2012convergence} have been developed to optimize the LASSO objective function. SPARCS however differs from LASSO as it does not consider a regularized objective function and does not require costly iterative optimization. Instead, it performs variable selection via thresholding 
the min-norm solution to the non-regularized OLS problem.

Offline implementation of the proposed SPARCS method can be compared with correlation learning, also called marginal regression, simple thresholding, and sure independence screening  \cite{genovese2009revisiting,genovese2012comparison,fan2008sure}, wherein the simple sample cross-correlation vector between the response variable and the predictor variables is thresholded. The theory developed in this paper also yields phase transitions for the familywise false discovery rate for these methods.

The SPARCS screening stage has some similarity to recently developed correlation screening and hub screening  in graphical models \cite{hero2011large,hero2012hub}. However, there are important and fundamental differences.  The methods in \cite{hero2011large,hero2012hub} screen for connectivity in the correlation graph, i.e., they only screen among the predictor variables $\{X_{1}, \ldots, X_{p}\}$.  SPARCS screens for the connections in the bi-partite graph between the response variable Y and the predictor variables $X_1,...,X_p$. Thus SPARCS is a supervised learning method that accounts for $Y$ while the methods of \cite{hero2011large,hero2012hub} are unsupervised methods.
%

SPARCS can also be compared to sequential sampling methods, originating in the pioneering work of \cite{wald1945sequential}. This work has continued in various directions such as sequential selection and ranking and adaptive sampling schemes \cite{bechhofer1968sequential,gupta1991sequential}. Recent advances include the many multi-stage adaptive support recovery methods that have been collectively called distilled sensing \cite{haupt2009compressive,haupt2011distilled,wei2013multistage,wei2013performance} in the compressive sensing literature. While bearing some similarities, our  SPARCS approach differs from distilled sensing (DS). Like SPARCS, DS performs initial stage thresholding in order to reduce the number of measured variables in the second stage. However, in distilled sensing the objective is to recover a few variables with high mean amplitudes from a larger set of initially  measured predictor variables. In contrast, SPARCS seeks to recover a few variables that are strongly  predictive of the response variable from a large number of initially measured predictor variables and the corresponding response variable. Furthermore, unlike in DS, in SPARCS the final predictor uses all the information on selected variables collected during both stages.

The paper is organized as follows. Section \ref{sec:SPARCS} provides a practical motivation for SPARCS from the perspective of an experimental design problem in biology. It introduces the under-determined multivariate regression problem and formally defines the two stages of the SPARCS algorithm. Section \ref{sec:Performance1} develops high dimensional  asymptotic analysis for screening and support recovery performance of SPARCS. Section \ref{sec:Performance1} also provides theory that specifies optimal sample allocation between the two stages of SPARCS. Section \ref{sec:Simulations} presents simulation comparisons and an application to symptom prediction from gene expression data.

\begin{figure} [ht]
\centering
\includegraphics[width=3.0in]{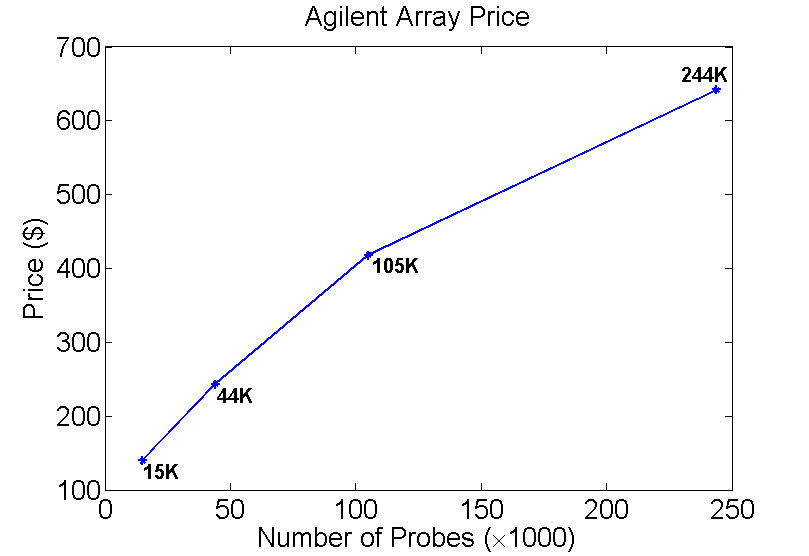}
\caption{\footnotesize Price of arrays as a function of the number of probes. The dots represent pricing per slide for Agilent Custom Microarrays G2509F, G2514F, G4503A, G4502A (May 2014). The cost increases as a function of probeset size. Source: BMC Genomics and RNA Profiling Core.} \label{fig:Agilent}
\end{figure}

\section{Two-stage SPARCS method for online sparse regression}
\label{sec:SPARCS}
In this section we motivate the two-stage SPARCS method for online sparse regression via an experimental design problem in biology. Moreover,  we formally define each stage of the two-stage SPARCS method.   

\subsection{Motivation and definition for SPARCS}
\label{subsec:SPARCSintro}
As a practical motivation for SPARCS consider the following sequential design problem that is relevant to applications where the cost of samples increases with the number $p$ of variables. This is often the case for example, in gene microarray experiments: a high throughput ``full genome" gene chip with $p=40,000$ gene probes can be significantly more costly than a smaller assay that tests fewer than $p=15,000$ gene probes (see Fig. \ref{fig:Agilent}).  In this situation a sensible cost-effective approach would be to use a two-stage procedure: first select a smaller number $l$ of variables on a few expensive high throughput samples and then construct the predictor on additional cheaper low throughput samples.

Motivated by the above practical example, we propose SPARCS as the following two-stage procedure. The first stage of SPARCS, also referred to as the SPARCS screening stage, performs variable selection and the second stage, also referred to as the SPARCS regression stage, constructs a predictor using the variables selected at the first stage. More specifically, assume that there are a total of $t$ samples $\{y_i,\bx_i\}_{i=1}^t$ available. During the first stage a number $n\leq t$ of these samples are assayed for all $p$ variables and during the second stage the rest of the $t-n$ samples are assayed for a subset of $l < p$ of the variables selected in the first stage. Variable selection at the SPARCS screening stage can be performed in one of two ways, (1) by screening the sample marginal cross-correlation between $Y$ and $\bX$, as in sure independence screening (SIS), or (2) by thresholding the solution to the generalized Ordinary Least Squares (OLS) problem, which we refer to as  predictive correlation screening (PCS). Subsequently,  the SPARCS regression stage uses standard OLS to design a $l$-variable predictor using all $t$ samples collected during both stages.



An asymptotic analysis (as the total number of samples $t \rightarrow \infty$) of the above two-stage predictor is undertaken in Sec. \ref{sec:Performance1} to obtain the optimal sample allocation for stage 1 and stage 2. Assuming that a sample of a single variable has unit cost and that the total available budget for all of the samples is $\mu$, the asymptotic analysis yields minimum Mean Squared Error (MSE) when $n$, $t$, $p$, and $k$ satisfy the budget constraint:
\begin{equation}
np + (t-n)k \leq \mu,
\label{eq:assaycostcond}
\end{equation}
where $k$ is the true number of active variables in the underlying linear model.  The condition in \eqref{eq:assaycostcond} is relevant in cases where there is a bound on the total sampling cost of the experiment and the cost of a sample increases linearly in its dimension $p$.

\subsection{SPARCS screening stage}
\label{subsec:Prelim}


 
We start out with some notations. Assume that $n$ i.i.d. paired realizations of $\bX=[X_{1},\ldots, X_{p}]$ and
$Y$ are available, where $\bX$ is a random vector of predictor variables and $Y$ is a scalar response variable to be predicted. We represent the $n \times p$ predictor data matrix as
$\mathbb X$ and the $n \times 1$ response data vector as $\mathbb Y$. 
The $p \times p$ sample covariance matrix $\bS^x$ for the rows of the data matrix $\mathbb X$ is defined as: \be \bS^x=\frac{1}{n-1} \sum_{i=1}^n
(\bx_{i}-\ol{\bx})^T(\bx_{i}-\ol{\bx}), \label{sampcov} \ee
where $\bx_{i}$ is the $i$-th row of data matrix $\mathbb X$, and
$\ol{\bx}$ is the vector average of all $n$ rows of $\mathbb X$. We also denote the sample variance of the elements of $\mathbb Y$ as $s^y$.

Consider the $n \times (p+1)$ concatenated matrix $\mathbb
W=[\mathbb X, \mathbb Y]$. The sample cross-covariance vector
$\bS^{xy}$ is defined as the upper right $p \times 1$ block of the
$(p+1) \times (p+1)$ sample covariance matrix obtained by
(\ref{sampcov}) using $\mathbb W$ as the data matrix instead of
$\mathbb X$. The $p \times p$ sample correlation matrix $\bR^{x}$ is defined as 
\begin{equation}
\bR^{x}= \bD_{\bS^x}^{-\frac{1}{2}} \bS^{x} \bD_{\bS^x}^{-\frac{1}{2}},
\label{eq:Rx}
\end{equation}
where $\bD_{\bA}$ represents a matrix that is obtained by zeroing out all but diagonal entries of $\bA$. Moreover, the $p \times 1$ sample cross-correlation vector $\bR^{xy}$ is defined as:
\begin{equation}
\bR^{xy}= \bD_{\bS^x}^{-\frac{1}{2}} \bS^{xy} (s^y)^{-\frac{1}{2}}.
\label{eq:Rxy}
\end{equation}

The SIS method for the SPARCS screening stage selects the desired number of variables, $l$, by picking the $l$ variables that have the largest absolute sample correlation with the response variable $Y$. Therefore, SIS performs support recovery by discovering the entries of $\bR^{xy}$ whose absolute value is larger than some threshold. 

Next we introduce the under-determined ordinary least squares (OLS) multivariate regression problem. 
Assume that $n < p$. We define the generalized Ordinary Least Squares (OLS) estimator of $Y$ given $\bX$ as the min-norm solution of the under-determined least squares regression problem 
\be \min_{\mathbf B^{xy} \in \mathbb{R}^p} \|\mathbb Y- \mathbb X \mathbf B^{xy}\|_F^2, \label{eq:OLS1}\ee
where $\| \bA \|_F$ represents the Frobenius norm of matrix $\bA$. The min-norm solution to \eqref{eq:OLS1} is the vector of regression coefficients
\be \mathbf B^{xy}=(\bS^x)^\dagger \bS^{xy}, \label{eq:OLS_solution}\ee
where $\bA^\dagger$ denotes the Moore-Penrose pseudo-inverse of the matrix $\bA$. If the $i$-th  entry of the regression coefficient vector $\mathbf B^{xy}$ is zero then the $i$-th predictor variable is not included in the OLS estimator. This is the main motivation for the PCS method for variable selection at the SPARCS screening stage.
More specifically, the PCS method selects  the $l$ entries of $\mathbf B^{xy}$ having the largest absolute values. Equivalently,  PCS performs support recovery by discovering the entries of the generalized OLS solution $\mathbf B^{xy}$ whose absolute value is larger than some threshold.


In Sec. \ref{subsec:recoveryguarantees} we will see that, under certain assumptions, SIS and PCS admit similar asymptotic support recovery guarantees. However, our experimental results in Sec. \ref{sec:Simulations}  show that for $n \ll p$, if SIS (or LASSO) is used instead of PCS in the SPARCS screening stage, 
the performance of the two-stage predictor suffers. This empirical observation suggests that pre-multiplication of $\bS^{xy}$ by the pseudo-inverse $(\bS^x)^{\dagger}$  instead of by the diagonal matrix  $\bD_{\bS^x}^{-1/2}$, can improve the performance of the SPARCS procedure.

\subsection{SPARCS regression stage}

In the second stage of SPARCS, a number $t-n$ of additional samples are collected for the $l<p$ variables found by the SPARCS screening stage.   Subsequently, a sparse OLS predictor of $Y$ is constructed using only the $l$ variables selected at the SPARCS screening stage. Specifically, the predictor coefficients are determined from all of the $t$ samples  according to 
\be
(\bS^x_{(l)})^{-1} \bS^{xy}_{(l)},
\ee
where $\bS^x_{(l)}$ and $\bS^{xy}_{(l)}$ are the $l \times l$ sample covariance matrix and the $l \times 1$ sample cross-covariance vector obtained for the set of $l$ variables selected by the SPARCS screening stage.

In Sec. \ref{sec:Performance1} we establish high dimensional statistical convergence rates for the two stage online SPARCS procedure and we obtain asymptotically optimal sample allocation proportions $n/t$ and $(t-n)/t$ for the first and second stage.

\section{Asymptotic analysis}
\label{sec:Performance1}

\subsection{Notations and assumptions}
In this section we introduce some additional notation and state the required assumptions for our asymptotic statistical analysis of SPARCS. 


 

The following notations are required for the 
propositions in this section. The surface area of the $(n-2)$-dimensional unit sphere $S_{n-2}$ in $\mathbb R^{n-1}$ is denoted by by $\ra_n$. In the sequel we often refer to a vector on $S_{n-2}$ as a \emph{unit norm} vector.

Our statistical analysis of SPARCS uses the U-score representations of the data. More specifically, there exist a $(n-1) \times p$ matrix $\mU^x$ with unit norm columns, and a $(n-1) \times 1$ unit norm vector $\mU^y$ such that the following representations hold \cite{hero2011large,hero2012hub}:
\be
\bR^{x} = (\mU^x)^T \mU^x,
\label{eq:Rx0}
\ee
and
\be
\bR^{xy} = (\mU^x)^T \mU^y.
\label{eq:Rxy0}
\ee
Specifically, the columns of the matrices $\mU^x$ and $\mU^y$ in the above representations are called U-scores. U-scores lie on the $(n-2)$-sphere $S_{n-2}$ in
$\Reals^{n-1}$ and are constructed by projecting away the component of the Z-scores that are orthogonal to the $n-1$ dimensional hyperplane $\{\bfu \in
\Reals^n: \mathbf1^T \bfu =0\}$. 
The sample correlation between $X_i$ and $X_j$ can be computed
using the inner product or the Euclidean distance between associated U-scores:
\be
\r^{x}_{ij}= (\bfU_i^x)^T \bfU_j^x=1- \frac{\|\bU_i^x -\bU_j^x\|_2^2}{2}.
\label{eq:Uscore_dist_rep}
\ee
Similarly, the sample correlation between $X_i$ and $Y$ can be computed as:
\be
\r^{xy}_{i}= (\bfU_i^x)^T \bfU^{y}=1- \frac{\|\bU_i^x -\bU^y\|_2^2}{2}.
\label{eq:Uscore_dist_rep}
\ee
More details about the U-scores representations can be found in  \cite{hero2011large,hero2012hub} and in the Appendix.

Assume that $\bU, \bV$ are two independent and
uniformly distributed random vectors on $S_{n-2}$. For a threshold
$\rho \in [0,1]$, let $r=\sqrt{2(1-\rho)}$. $P_0(\rho,n)$ is then defined as the
probability that either $\|\bU-\bV\|_2 \leq r$ or $\|\bU+\bV\|_2 \leq r$. $P_0(\rho,n)$ can be computed using the formula for the area of spherical caps on $S_{n-2}$ (cf. \cite{li2011concise}):
\be
P_0 = I_{1-\rho^2}((n-2)/2,1/2),
\ee
in which $I_x(a,b)$ is the regularized incomplete beta function.

$S \subseteq \{1,\ldots,p\}$ denotes the set of indices of the variables selected by the SPARCS screening stage. Moreover, $l$ refers to the number of variables selected at the SPARCS screning stage, i.e., $|S| = l$.



For the asymptotic analysis we assume that the response $Y$ is generated from the following statistical model:
\be Y = a_{i_1} X_{i_1} + a_{i_2} X_{i_2} + \cdots + a_{i_k} X_{i_k} + N , \label{eq:LinearModel} \ee where $\pi_0 = \{ i_1,\cdots, i_k \}$ is a set of distinct indices in $\{1,\ldots,p \}$, $\bX = [X_1, X_2,\cdots,X_p]$ is the vector of predictors, $Y$ is the response variable, and $N$ is a noise variable. $X_{i_1},\cdots,X_{i_k}$ are called active variables and the remaining $p-k$ variables are called inactive variables. In the sequel, we refer to the set $\pi_0$ as the support set, and $|\pi_0| = k$ denotes the number of active variables. 

Unless otherwise specified, throughout this paper we consider random data matrices $\mathbb X$ that satisfy the following: for every $\epsilon > 0$ there exist a constant $C > 0$ such that the following concentration property holds:
\begin{equation}
\mathbb P \big(~ \| \bD_{\bS^x} - \bD_{\mathbf \Sigma_x} \| > \epsilon ~ \big) < \exp(-Cn), \label{CProperty}
\end{equation}
in which $\|\bA \|$ is the operator norm of $\bA$, $\bS^x$ is the sample covariance matrix defined in \eqref{sampcov}, and $\mathbf \Sigma_x$ is the population covariance matrix.  Property \eqref{CProperty} is similar to, but weaker than, the concentration property introduced in \cite{fan2008sure} as it only implies bounds on the joint convergence rate of the diagonal entries of the sample covariance matrix (i.e., sample variances of the predictor variables $X_1,\cdots,X_p$) and does not imply bounds on the convergence of the off-diagonal entries of the sample covariance matrix (i.e., sample cross covariances of the predictor variables $X_1,\cdots,X_p$). It is known that the concentration property \eqref{CProperty} holds when the predictors $\bX$ follow a $p$-variate distribution with sub-Gaussian tails \cite{koltchinskii2014concentration}. It is  worth mentioning that the concentration property \eqref{CProperty} is also satisfied when the linear model \eqref{eq:LinearModel} is assumed on the standardized observations for which $\bD_{\bS^{x}} = \bD_{\mathbf \Sigma_x} = \bI_p $.

In our asymptotic analysis of SPARCS we make  the following additional  assumptions on the linear model \eqref{eq:LinearModel}, which are comparable or weaker than assumptions made in other studies  \cite{obozinski2011support,fan2008sure,candes2007sparsity,tropp2007signal,carin2011coherence}.

\begin{assumption}  The $n \times p$ data matrix $\mX$ follows a multivariate elliptically contoured distribution with mean $\boldsymbol \mu_x$ and $p \times p$ dispersion matrix $\mathbf{\Sigma}_x$,  i.e. the probability density function (pdf) is of the form $f_{\mathbb X}(\mathbb X) = g \Bigl( {\mathrm tr} \left( (\mathbb X- \mathbf 1 \mathbf \mu_x^T) \Sigma_x^{-1}(\mathbb X- \mathbf 1 \mathbf \mu_x^T)^T \right) \Bigr)$, where $g$ is a non-negative function and ${\mathrm tr(\mathbf A)}$ is the trace of $\mathbf A$. Moreover, the density function $f_{\mathbb X}(.)$ is bounded and differentiable.
\label{Asmp:elliptical}
\end{assumption}


\begin{assumption}
Let $\rho_{yi}$ represent the true correlation coefficient between response variable $Y$ and predictor variable $X_i$.
The quantity
\be \label{eq:rhomin}
\rho_{\min} = \min_{i \in \pi_0, j \in \{1,\cdots,p \} \backslash \pi_0}\{|\rho_{yi}| - |\rho_{yj}|\}, \ee
is strictly positive and independent of $p$.
 \label{Asmp:rhomin} \end{assumption}

\begin{assumption} The $(n-1) \times p$ matrix of U-scores satisfies (with prob. 1): \be \frac{n-1}{p}\mU^x (\mU^x)^T = \bI_{n-1} + \textbf o(1),~~~~~~~\text{as}~~p \rightarrow \infty,  \ee in which $\textbf o(1)$ is a $(n-1) \times (n-1)$ matrix whose entries are $o(1)$.\label{Asmp:coherency} \end{assumption}



Assumption \ref{Asmp:elliptical} is weaker than the Gaussian assumption commonly used in compressive sensing \cite{haupt2011distilled,baraniuk2010model} and, unlike standard sub-Gaussian assumptions commonly used in high dimensional data analysis \cite{buhlmann2011statistics}, allows for heavy tails. 
Assumption \ref{Asmp:rhomin} is a common assumption that one finds in performance analysis of support recovery algorithms (cf.  \cite{obozinski2011support,fan2008sure}). In particular, Assumption \ref{Asmp:rhomin} can be compared to the conditions on the sparsity-overlap function in \cite{obozinski2011support} which impose assumptions on the population covariance matrix in relation to the true regression coefficients. Assumption \ref{Asmp:rhomin} can also be compared to Condition 3 introduced in \cite{fan2008sure} that  imposes lower bounds on the magnitudes of the true regression coefficients as well as  on the true correlation coefficients between predictors and the response. Assumption \ref{Asmp:coherency} can be related to assumptions (A1)-(A3) in \cite{obozinski2011support} in the sense that they both lead to regularity conditions on the entries and the eigenspectrum of the correlation matrix. Assumption \ref{Asmp:coherency} is also similar to the concentration property introduced in \cite{fan2008sure} as they both yield regularity conditions on the inner products of the rows of the data matrix. Moreover, Assumption \ref{Asmp:coherency} can also be considered as an incoherence-type condition on the U-scores, similar to the incoherence conditions on the design matrix assumed in the compressive sensing literature \cite{candes2007sparsity,tropp2007signal,carin2011coherence}. It is worth mentioning that a special case in which Assumption \ref{Asmp:coherency} is satisfied is the orthogonal setting where $\mathbb{X} \mathbb{X}^T /n = \bI_n$. 

Lemma \ref{Lemma:SigmaClass} below specifies a class of $p \times p$ correlation matrices $\mathbf{\Omega}_x$ for which Assumption \ref{Asmp:coherency} is satisfied.

\begin{lemmas}
\label{Lemma:BlockSparse}
Assume that the population correlation matrix $\mathbf{\Omega}_x = \bD_{\mathbf\Sigma_x}^{-1/2} \mathbf\Sigma_x \bD_{\mathbf\Sigma_x}^{-1/2}$ is of the following weakly block-sparse form \be \mathbf{\Omega}_x = \mathbf{\Omega}_{bs} + \mathbf \Omega_{e}, \label{eq:bsplustoeplitz}
\ee
in which $\mathbf{\Omega}_{bs}$ is a $p \times p$ block-sparse matrix of degree $d_x$ (i.e., by re-arranging rows and columns of $\mathbf{\Omega}_{bs}$ all non-zero off-diagonal entries can be collected in a $d_x \times d_x$ block), and $\mathbf \Omega_{e} = [\omega_{ij}]_{1\leq i,j \leq p}$ is a $p \times p$ matrix such that $\omega_{ij} = O\left(f(|i-j|)\right)$ for some function $f(.)$ with $\lim_{t \rightarrow \infty} f(t) = 0$. 
If $d_x = o(p)$,
then Assumption \ref{Asmp:coherency} holds.
\label{Lemma:SigmaClass}
\end{lemmas}
\noindent{\it Proof of Lemma \ref{Lemma:SigmaClass}:} See Appendix. \qed

Note that Lemma \ref{Lemma:SigmaClass} is essentially a result of the application of the law of large numbers to the inner product of the rows of the U-score matrix $\mU^x$. More specifically, due to specific decomposition \eqref{eq:bsplustoeplitz} for the correlation matrix $\mathbf{\Omega}_{x}$, as $p \rightarrow \infty$, the inner product of two different rows of $\mU^x$ converges to 0, as the proportion of the terms that are obtained by multiplication of significantly correlated variables converges to zero.

\subsection{High dimensional asymptotic analysis for screening}
\label{subsec:pvalues}

 
In this section, we establish a Poisson-like limit theorem for the number of variables that pass the SPARCS screening stage as $p\rightarrow\infty$ for fixed $n$. This yields a Poisson approximation to the probability of false discoveries that is accurate for small $n$ and large $p$. The Poisson-like limit theorem also specifies a phase transition threshold for the false discovery probability. 



 
Lemma below states that the PCS method can be interpreted as a method for discovering the non-zero entries of a $p \times 1$ vector with a special representation, by thresholding the entries at some threshold $\rho$. It is worth noting that a similar result also holds true for SIS without Assumption \ref{Asmp:coherency}.

\begin{lemmas}
Under Assumptions \ref{Asmp:elliptical} and \ref{Asmp:coherency}, the PCS algorithm for support recovery is asymptotically equivalent to  thresholding the entries of a $p \times 1$ vector $\mathbf \Phi^{xy}$ which admits the following representation: 
\be
\mathbf \Phi^{xy} = (\mZ^x)^T \mZ^y,
\label{eq:Phixy}
\ee
in which $\mZ^x$ is a $(n-1) \times p$ matrix whose columns are unit norm vectors, and $\mZ^y$ is a $(n-1) \times 1$ unit norm vector.
\label{lemma:Phixy}
\end{lemmas}
\noindent{\it Proof of Lemma \ref{lemma:Phixy}:} See Appendix. \qed

For a threshold $\rho \in [0,1]$, let $N^{xy}_{\rho}$ denote the number of entries of a $p \times 1$ vector of the form \eqref{eq:Phixy} whose magnitude is at least $\rho$.
The following proposition gives an asymptotic expression for the expected number of discoveries $\mathbb E[N^{xy}_{\rho}]$, for fixed $n$, as $p \rightarrow \infty$ and $\rho \rightarrow 1$. It also
states that under certain assumptions, the probability of having at
least one discovery converges to a given limit. This limit is
equal to the probability that a certain Poisson random variable $N^*$ with rate equal to $\lim_{p\rightarrow \infty, \rho \rightarrow 1} \mathbb E[N^{xy}_{\rho}]$ satisfies: $N^*>0$. The following proposition does not need the concentration property \eqref{CProperty} to hold.


\begin{propositions}
Consider the linear model \eqref{eq:LinearModel}. Let $\{\rho_p\}_p$ be a sequence of threshold values in $[0,1]$ such
that $\rho_p \rightarrow 1$ as $p\rightarrow \infty$ and
$p(1-\rho_p^2)^{(n-2)/2} \rightarrow
e_{n}$. Under the Assumptions \ref{Asmp:elliptical} and \ref{Asmp:coherency}, if the number of active variables $k$ grows at a rate slower than $p$, i.e., $k = o(p)$, then for the number of discoveries $N^{xy}_{\rho_p}$ we have:
\be
\lim_{p\rightarrow \infty} \mathbb E[N^{xy}_{\rho_p}]= \lim_{p\rightarrow \infty}
\xi_{p,n,\rho_p} = \zeta_{n}, \label{xidef_p2}
  \ee where $
\xi_{p,n,\rho_p}=p P_0(\rho,n)$ and $\zeta_{n}= e_{n}
\ra_n/(n-2)$. Moreover:
\be \lim_{p\rightarrow \infty} \mathbb{P}(N_{\rho_p}^{xy}>0) = 1-\exp(\zeta_{n}).
\label{eq:Poissonconvcross_p2}
\ee
\label{Prop2}
\end{propositions}
\noindent{\it Proof of Proposition \ref{Prop2}:} See Appendix. \qed

Note also that Prop. \ref{Prop2} can be generalized to the case where Assumption \ref{Asmp:coherency} is not required. However when Assumption \ref{Asmp:coherency} is removed the asymptotic rates for $\mathbb E[N^{xy}_{\rho_p}]$ and $\mathbb{P}(N_{\rho_p}^{xy}>0)$ depend on the underlying distribution of the data. Such a generalization of Prop. \ref{Prop2} is given in the Appendix.

\begin{figure}
\begin{center}
\begin{tikzpicture}[thin]

\node at (-0.1,-0.5) {$\textbf{Part x}$};
\node at (0,-1) {$X_1$};
\node at (0,-1.5) {$X_2$};
\node at (0,-2.5) {$X_i$};
\node at (0,-3.5) {$X_p$};
\node at (2.6,-0.75) {$\textbf{Part y}$};
\node at (2.75,-2.25) {$Y$};

\node at (0.35,-1) [matched]{};
\node at (0.35,-1.5) [matched]{};
\node at (0.35,-2.5) [matched]{};
\node at (0.35,-3.5) [matched]{};
\node at (2.4,-2.25) [matched]{};

\draw [dotted] (0.35,-1) -- (1,-1.4);
\draw [dotted](0.35,-1.5) -- (1,-1.75);
\draw [dotted] (0.35,-3.5) -- (1,-3.1);

\draw [dotted](2.4,-2.25) -- (1.65,-1.85);
\draw [dotted](2.4,-2.25) -- (1.65,-2.00);
\draw [dotted](2.4,-2.3) -- (1.65,-2.75);

\draw (0.35,-2.5) -- (2.4,-2.25);
\draw[loosely dotted]  (0.35,-1.8) -- (0.35,-2.2);
\draw[loosely dotted]  (0.35,-2.7) -- (0.35,-3.3);

\end{tikzpicture}

\caption{\footnotesize The first stage of SPARCS is equivalent to discovering the non-zero entries of the $p \times 1$ vector $\mathbf \Phi^{xy}$ in \eqref{eq:Phixy} to find variables $X_i$ that are most predictive of the response  $Y$. This is equivalent to finding sparsity in a bipartite graph ${\mathcal G}_{\rho}(\mathbf \Phi^{xy})$ with parts $x$ and $y$ which have vertices $\{X_1,\ldots,X_p\}$ and $Y$, respectively. For $1 \leq i \leq p$, vertex $X_i$ in part $x$ is connected to vertex $Y$ in part $y$ if
$|\phi_{i}^{xy}| > \rho$.} \label{fig:bipartite_graph}
\end{center}
\end{figure}
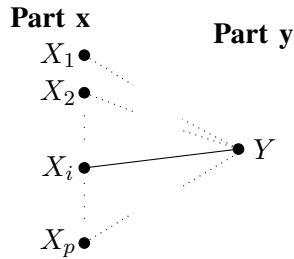

Proposition \ref{Prop2} plays an important role
in identifying phase transitions and in approximating $p$-values associated with individual predictor variables. More specifically, under the assumptions of Prop. \ref{Prop2}:
\be \mathbb P(N_{\rho_p}^{xy}>0) \rightarrow
1-\exp(-\xi_{p,n,\rho_p}) ~~\text{as} ~~ p \rightarrow
\infty.
 \ee
The above limit provides an approach for calculating approximate p-values in the setting where the dimension $p$ is very large.
For a threshold
$\rho \in [0,1]$ define ${\mathcal G}_{\rho}(\mathbf \Phi^{xy})$ as the undirected bipartite graph (Fig. \ref{fig:bipartite_graph}) with parts labeled
$x$ and $y$, and vertices $\{X_1,X_2,...,X_p \}$ in part $x$ and
$Y$ in part $y$. For $1 \leq i \leq p$, vertices $X_i$ and
$Y$ are connected if $|\phi_{i}^{xy}|
>\rho$, where $\phi_{i}^{xy}$ is the $i$-th entry of $\mathbf \Phi^{xy}$ defined in
\eqref{eq:Phixy}. Denote by $d_i^x$ the degree of vertex $X_i$ in ${\mathcal
G}_{\rho}(\mathbf \Phi^{xy})$. Note that $d_i^x \in \{0,1\}$. For each $1 \leq i \leq p$, denote by
$\rho(i)$ the maximum value of the threshold $\rho$
for which $d^x_i=1$ in ${\mathcal G}_\rho(\mathbf \Phi^{xy})$. By this definition, we have $\rho(i) = |\phi_{i}^{xy}|$. 
Using Prop. \ref{Prop2} the $p$-value associated with predictor variable $X_i$ can now be approximated as: \be pv(i) \approx
1-\exp(-\xi_{p,n,\rho(i)}). \label{eq:p-vals} \ee 
Similar to the result in \cite{hero2011large,hero2012hub}, there is a
phase transition in the $p$-values as a function of the threshold $\rho$. More
exactly, there is a critical threshold $\rho_{c}$ such
that if $\rho > \rho_{c}$, the average
number $\mathbb E[N_{\rho}^{xy}]$ of discoveries abruptly decreases to $0$ and if $\rho < \rho_{c}$ the average
number of discoveries abruptly increases to $p$. Motivated by this, we define the critical threshold $\rho_c$ as the threshold that satisfies the equation $\partial \mathbb E[N_{\rho}^{xy}]/\partial \rho = -p$. Using \eqref{xidef_p2}, the value of the critical threshold can be approximated as:
\be
\rho_{c}=\sqrt{1-(\ra_n p)^{-2/(n-4)}}.
\label{critical} \ee 
Note that the expression given in \eqref{critical} bears resemblance to the expression $(3.14)$ in \cite{hero2011large}. 
Expression (\ref{critical}) is useful in choosing the screening threshold $\rho$. 
Selecting $\rho$ slightly greater than
$\rho_{c}$ will prevent the bipartite
graph ${\mathcal G}_{\rho}(\mathbf \Phi^{xy})$ from having an overwhelming number of edges.

\subsection{High dimensional asymptotic analysis for support recovery}
\label{subsec:recoveryguarantees}

In this section we give theoretical upper bounds on the Family-Wise Error Rate (FWER) when performing variable selection in SPARCS screening stage.

Propositions \ref{Prop:UpperBoundAlg2} and \ref{Prop:UpperBoundAlg1} give upper bounds on the probability of selection error for the SPARCS screening stage by thresholding the vector $\mathbf R^{xy}$ (i.e. using SIS), or the vector $\mathbf B^{xy}$ (i.e. using PCS), respectively.

\begin{propositions}
\label{Prop:UpperBoundAlg2}
Let $S$ denote the support set selected using SIS and let $l = |S|$ be the size of this support.  Under Assumptions \ref{Asmp:elliptical} and \ref{Asmp:rhomin}, if $n \geq \Theta(\log p)$ then for any $l \geq k$, 
SIS recovers the support $\pi_0$, with probability at least $1-1/p$, i.e. \be \mathbb{P}\left( \pi_0 \subseteq S \right) \geq 1-1/p. \ee
\end{propositions} 
\noindent{\it Proof of Proposition \ref{Prop:UpperBoundAlg2}}: See Appendix. \qed



\begin{propositions}
\label{Prop:UpperBoundAlg1}
Let $S$ denote the support set selected using PCS and let $l = |S|$ be the size of this support.  Under Assumptions \ref{Asmp:elliptical}-\ref{Asmp:coherency}, if $n \geq \Theta(\log p)$ then for any $l \geq k$, 
PCS recovers the support $\pi_0$, with probability at least $1-1/p$, i.e. \be \mathbb{P}\left( \pi_0 \subseteq S \right) \geq 1-1/p.  \ee
\end{propositions}
\noindent{\it Proof of Proposition \ref{Prop:UpperBoundAlg1}:}  See Appendix.

\qed


The constant in $\Theta(\log p)$ of Prop. \ref{Prop:UpperBoundAlg2} and Prop. \ref{Prop:UpperBoundAlg1} is increasing in $\rho_{\min}$. It is shown in the proof of the propositions that $12/\rho_{\min}$ is an upper bound for the constant in $\Theta(\log p)$. Note that  the above propositions on support recovery allow all types of non-zero correlations (i.e., correlations between active variables, correlations between inactive variables, and correlations between active and inactive variables) as long as the corresponding assumptions 
are satisfied. 

Propositions \ref{Prop:UpperBoundAlg2} and  \ref{Prop:UpperBoundAlg1} can be compared to Thm. $2$ in \cite{obozinski2011support} and Thm. $1$ in \cite{fan2008sure} for recovering the support set $\pi_0$. 
More specifically, Thm. $2$ in \cite{obozinski2011support} asserts a similar result as in Prop. \ref{Prop:UpperBoundAlg2} and Prop. \ref{Prop:UpperBoundAlg1} for support recovery via minimizing a LASSO-type objective function. Also Thm. $1$ in \cite{fan2008sure} asserts that if $n = \Theta((\log p)^{\alpha})$ for some $\alpha>1$, SIS recovers the true support with probability no less than $1-1/p$. 
Note also that Prop. \ref{Prop:UpperBoundAlg2} and Prop. \ref{Prop:UpperBoundAlg1} state stronger results than the similar results proven in \cite{fan2008sure} and in \cite{obozinski2011support}, respectively, in the sense that the support recovery guarantees presented in \cite{fan2008sure,obozinski2011support} are proven for the class of multivariate Gaussian distributions whereas Prop. \ref{Prop:UpperBoundAlg2} and Prop. \ref{Prop:UpperBoundAlg1} consider the larger class of multivariate elliptically contoured distributions. These distributions accommodate heavy tails.

\subsection{High dimensional asymptotic analysis for prediction}
\label{subsec:OptimalSample}
The following proposition states the optimal sample allocation rule for the two-stage SPARCS predictor, in order to minimize the expected MSE as $t \rightarrow \infty$. 

\begin{propositions}
\label{Prop:UpperBound2}
The optimal sample allocation rule for the SPARCS online procedure introduced in Sec. \ref{sec:SPARCS} under the cost condition \eqref{eq:assaycostcond} is 
\be n = \left\{\begin{array}{cc} O(\log t), & c(p-k) \log t+k t \leq \mu\\
0, & o.w.
\end{array} \right.
\ee
where $c$ is a positive constant that is independent of $p$.
\end{propositions}
\noindent{\it Proof of Proposition \ref{Prop:UpperBound2}:} See Appendix. \qed

\begin{figure}[h!]
\centering
\subfigure {\includegraphics[width=2.8in]{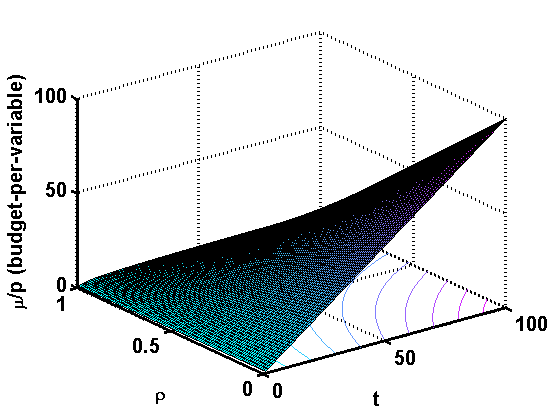}}
\subfigure {\includegraphics[width=2.8in]{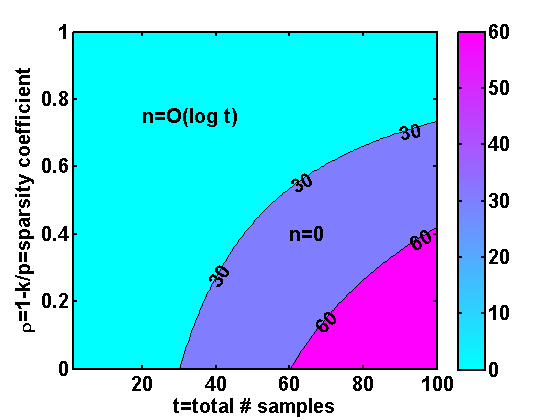}}
\caption{\footnotesize (Left) Surface $\mu/p=c \rho \log t+(1-\rho) t$, for $c=1$. (Right) Contours indicating optimal allocation regions for $\mu/p=30$ and $\mu/p=60$ ($\rho=1-k/p$). As the coefficient $c$ increases, the surface $c \rho \log t+(1-\rho) t$ moves upward and the regions corresponding to $n=O(\log t)$ and $n=0$, become smaller and larger, respectively.}
\label{fig:budget}
\end{figure}

The constant $c$ above is an increasing function of the quantity $\rho_{\min}$ defined in \eqref{eq:rhomin}. Proposition \ref{Prop:UpperBound2} asserts that for a generous budget ($\mu$ large) the optimal first stage sampling allocation is  $O(\log t)$. However, when the budget is tight it is better to skip stage 1 ($n=0$). Figure \ref{fig:budget} illustrates the allocation region (for $c = 1$) as a function of the sparsity coefficient $\rho=1-k/p$. Note that Prop. \ref{Prop:UpperBound2} is generally true for any two-stage predictor which at the first stage, uses a support recovery method that satisfies the performance bound proposed by Prop. \ref{Prop:UpperBoundAlg2} or Prop. \ref{Prop:UpperBoundAlg1}, and at the second stage uses OLS. 

\section{Numerical comparisons}
 \label{sec:Simulations}
\label{sec:Simulations}

We now present experimental results which demonstrate the performance of SPARCS  when applied to both synthetic and real world data. Throughout this section we refer to the SPARCS predictors which use SIS or PCS at the first stage as SIS-SPARCS or PCS-SPARCS, respectively.

\textit{a) Efficiency of SPARCS screening stage.}
We illustrate the performance of the SPARCS screening stage (i.e., the first stage of the SPARCS predictor) using SIS or PCS and compare these to LASSO \cite{tibshirani1996regression,genovese2012comparison}. 

In the first set of simulations we generated an $n \times
p$ data matrix $\mX$ with independent rows, each of which is
drawn from a $p$-dimensional multivariate normal distribution with mean $\mathbf{0}$ and block-sparse covariance matrix satisfying \eqref{eq:bsplustoeplitz}. 
The $p \times 1$ coefficient vector
$\ba$ is then generated such that exactly $100$ entries of $\ba \in \mathbb{R}^p$ are active. Each active entry of $\ba$ is an independent draw from $\mathcal{N}(0,1)$ distribution, and each inactive entry of $\ba$ is zero.
Finally, a synthetic response vector $\mY$ is generated
by a simple linear model \be \mY =  \mX \ba + \mN, \label{lm} \ee 
where $\mN$ is $n \times 1$ noise vector whose entries are i.i.d.
$\mathcal{N}(0,0.05)$. The importance of a variable is measured
by the magnitude of the corresponding entry of $\ba$. 

\begin{figure}[h!] 
\centering
\includegraphics[width=3.0in]{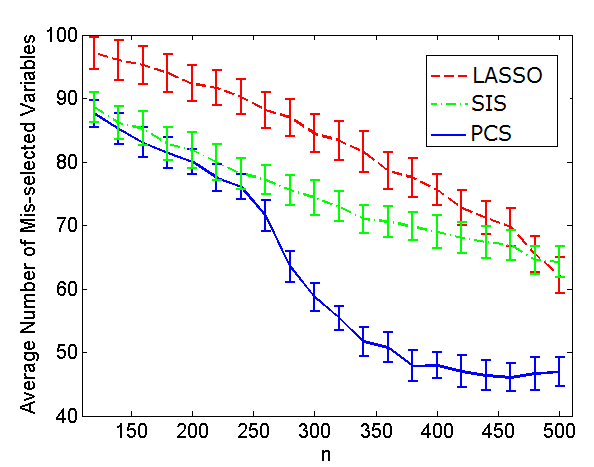}
\caption{\footnotesize Average number of mis-selected variables. Active set implementation of LASSO (red-dashed)
vs. SIS (green-dashed) vs. PCS (solid), $p=10000$. The data is generated via model \eqref{lm}. The regularization parameter of LASSO is set using 2-fold cross validation. It is evident that PCS has a lower miss-selection error compared to SIS and LASSO.}
\label{fig:LASSO100}
\end{figure}

We implemented LASSO on the above data set using an active set type algorithm - asserted to be one the fastest methods for solving LASSO \cite{kim2010fast}. In all of our implementations of LASSO, the regularization parameter is tuned to minimize prediction MSE using 2-fold cross validation. 
To illustrate SPARCS screening stage for a truly high dimensional example, we set $p=10000$ and
compared SIS and PCS methods with LASSO, for a small number of samples. Figure
\ref{fig:LASSO100} shows the results of this simulation over
an average of $400$ independent experiments for each value of $n$. 
As we see for small number of samples, PCS and SIS methods perform significantly better in selecting the important predictor variables. Moreover, the advantage of the extra pseudo-inverse factor used for variable selection in PCS as compared to SIS is evident in Fig. \ref{fig:LASSO100}.

\begin{figure}[h!]
\centering
\subfigure{\includegraphics[width=2.9in]{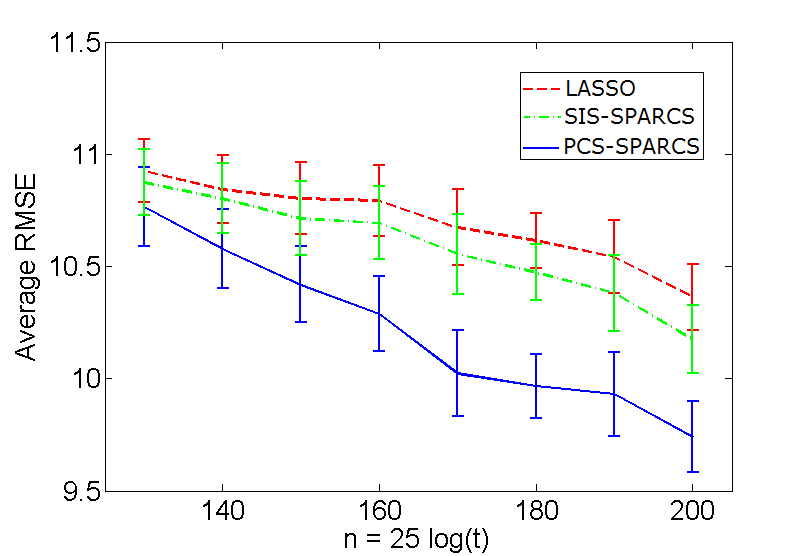}}
\subfigure{\includegraphics[width=2.9in]{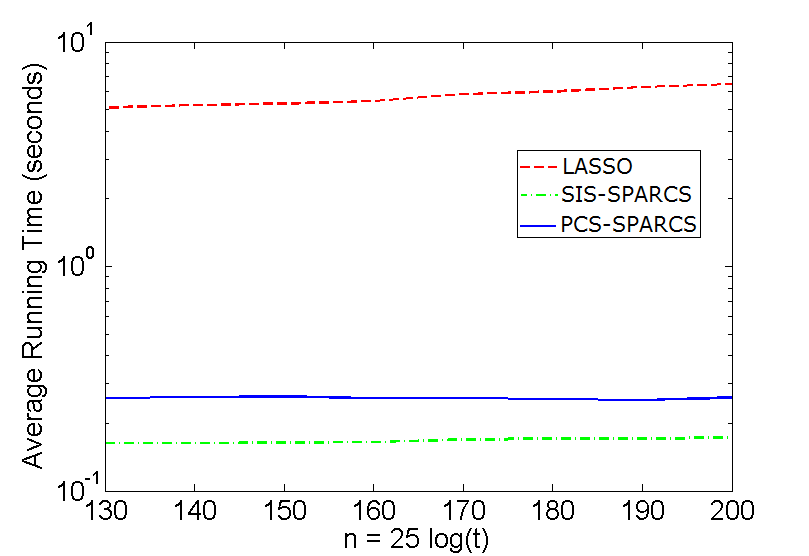}}
\caption{\footnotesize (Left) Prediction RMSE for the two-stage predictor when $n = 25 \log t$ samples are used  for screening at the first stage and all $t$ samples are used for computing the OLS estimator coefficients at the second stage. The solid plot shows the RMSE for PCS-SPARCS while the green and red dashed plots show the RMSE for SIS-SPARCS and LASSO, respectively. Here, $p=10000$. The Oracle OLS (not shown), which is the OLS predictor constructed on the true support set, has average RMSE performance that is a factor of 2 lower than the curves shown in the figure. This is due to the relatively small sample size available to these algorithms. (Right) Average running time as a function of $n$ for the experiment of the plot on the left. It is evident that due to lower computational complexity, SIS-SPARCS and PCS-SPARCS run an order of magnitude faster than LASSO.}
\label{fig:p100002n_MSE}
\end{figure}

\textit{b) Efficiency of the SPARCS predictor.} To test the efficiency of the proposed SPARCS predictor, a total of $t$ samples are generated using the linear model \eqref{lm} from which $n = 25 \log t$ are used for the task of variable selection at the first stage. All $t$ samples are then used to
compute the OLS estimator restricted to the selected variables. We chose $t$ such that $n = (130:10:200)$. The performance is evaluated by the empirical Root Mean Squared Error \be \text{RMSE} = \sqrt{\sum_{i=1}^{m} (y_i-\hat{y}_i)^2 / m}, \ee
where $m$ is the number of simulation trials. Similar to the previous experiment, exactly $100$ entries of $\ba$ are active and the predictor variables follow a multivariate normal distribution with mean $\mathbf{0}$ and block-sparse covariance matrix. 
Figure \ref{fig:p100002n_MSE} shows
the result of this simulation for $p=10000$, in terms of performance (left) and running time (right).
Each point on these plots is an average of $1000$ independent
experiments. Observe that in this low sample regime, when LASSO or SIS are used instead of PCS in the first stage, the performance suffers. More specifically we observe that the RMSE of the PCS-SPARCS predictor is uniformly lower than the SIS-SPARCS predictor or the two-stage predictor that uses LASSO in the first stage. 
 Table \ref{table:ttest} shows the $p$-values of one-sided paired t-tests testing for differences between the RMSE for PCS-SPARCS and SIS-SPARCS (LASSO) for several different values of $n$. These results show the high statistical significances of these RMSE differences.

\begin{figure} [h!]
\centering
\subfigure{\includegraphics[width=2.9in]{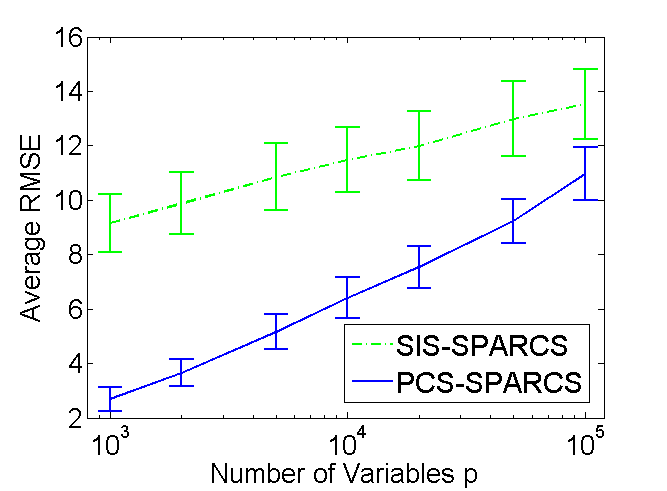}}
\subfigure{\includegraphics[width=2.9in]{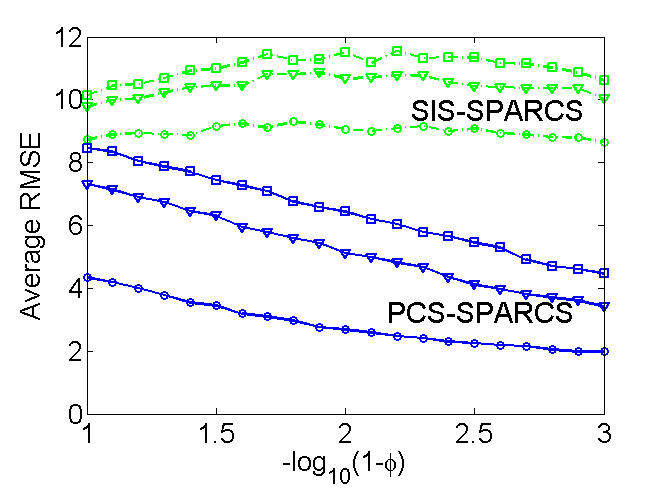}}
\caption{\footnotesize (Left) Prediction RMSE for the two-stage predictor when $n=500$ samples are used at the first stage, and a total of $t=2000$ samples are used at the second stage. The number of variables varies from $p=1000$ to $p=100000$. In this experiment, inactive variables are generated via realizations of an Auto-Regressive process of the form \eqref{ARprocess} with $\phi = 0.99$ ($-\log_{10}(1-\phi)=2$). The solid and dashed plots show the RMSE for PCS-SPARCS and SIS-SPARCS, respectively. The plots show the advantage of using PCS instead of SIS at the SPARCS screening stage. (Right) Prediction RMSE as function of the multicollinearity coefficient $-\log_{10}(1-\phi)$ for $p=[1000, 5000, 10000]$. For both PCS-SPARCS (solid) and SIS-SPARCS (dashed) predictors, the plots with square, triangle and circle markers correspond to $p=10000, p=5000$ and $p=1000$, respectively. These plots show that the PCS-SPARCS predictor uniformly outperforms the SIS-SPARCS predictor. Observe also that as the multicollinearity coefficient $-\log_{10}(1-\phi)$ increases the performance of the PCS-SPARCS predictor improves.}
\label{fig:SPARCvsSIS}
\end{figure}

\begin{table*}[t]
\setlength{\tabcolsep}{1pt}
\begin{center}
\scriptsize
\begin{tabular}{|c|c|c|c|c|c|c|c|c|c|}  \hline
$n$ & $130$ & $140$ & $150$ & $160$ & $170$ & $180$ & $190$ & $200$  \\ \hline
\tiny{PCS-SPARCS vs. SIS-SPARCS} & $7.7 \times 10^{-3}$ & $6.7\times 10^{-09}$& $3.2\times 10^{-11}$& $2.4\times 10^{-22}$& $7.8\times 10^{-29}$ & $8.1\times 10^{-36}$ & $9.2\times 10^{-42}$ & $5.3\times 10^{-46}$ \\ \hline
PCS-SPARCS vs. LASSO & $3.1 \times 10^{-4}$ & $8.0\times 10^{-10}$ & $7.2\times 10^{-14}$ & $3.0\times 10^{-25}$ & $1.8\times 10^{-30}$ & $5.6\times 10^{-39}$ & $1.1\times 10^{-42}$ & $6.5\times 10^{-48}$ \\ \hline
\end{tabular}
\caption{\footnotesize $p$-values of the one-sided paired t-test for testing the null hypothesis $\mathcal{H}_0$: PCS-SPARCS and SIS-SPARCS (LASSO) have the same average prediction RMSE in the experiment corresponding to Fig \ref{fig:p100002n_MSE}. Small $p$-values suggest that PCS-SPARCS significantly outperforms the others.}
\label{table:ttest}
\end{center}
\end{table*}
To further indicate the advantage of the PCS-SPARCS predictor compared to the SIS-SPARCS predictor, we performed simulations in which the number of samples used at the first stage, $n=500$, and the number of samples used at the second stage, $t=2000$, are fixed while the number of variables $p$  increases from $p=1000$ to $p=100000$. Moreover, exactly $100$ entries of the coefficient vector $\ba$ are active. Similar to the previous experiments, samples are generated using the linear model \eqref{lm}. However, in order to generate a data set with high multicollinearity, a scenario   that is likely to happen in high dimensional data sets (see \cite{rajaratnam2014deterministic} and the references therein), here the inactive variables are consecutive samples of an Auto-Regressive (AR) process of the form:
\be
&&W(1) = \epsilon(1), \nonumber \\
&&W(i) = \phi W(i-1) + \epsilon(i),~~ i = 2,\ldots,p-100,
\label{ARprocess}
\ee
in which $\epsilon(i)$'s are independent draws of $\mathcal{N}(0,1)$. The result of this experiment for $\phi=0.99$ is shown in Fig. \ref{fig:SPARCvsSIS} (left). The average RMSE values are computed using $1000$ independent experiments. The advantage of using PCS-SPARCS over SIS-SPARCS is evident in Fig. \ref{fig:SPARCvsSIS} (left). Note that as the number of variables $p$ becomes significantly larger than the number of samples $n$, the performance of both of the predictors converge to the performance of a random selection and estimation scheme in which variables are selected at random in the first stage. 

Furthermore, to analyze the performance of PCS-SPARCS and SIS-SPARCS for different levels of multicollinearity in the data, we performed similar experiments for $p=[1000, 5000, 10000]$ as the value of $\phi$ increases from $0.9$ to $0.999$. Figure \ref{fig:SPARCvsSIS} (right) shows the result of this simulation. Each point on these plots is the average of $500$ independent experiments. It is evident that similar to the previous experiment, the PCS-SPARCS predictor outperforms the SIS-SPARCS predictor. An interesting observation in Fig \ref{fig:SPARCvsSIS} (right) is that as the multicollinearity coefficient $-\log_{10}(1-\phi)$ increases the performance of the PCS-SPARCS predictor improves.

\textit{c) Estimation of FWER using Monte Carlo simulation.}
We set $p = 1000, k=10, n=[100,200,\ldots,1000]$ and using Monte Carlo simulation, we computed the probability of support recovery error for the PCS method. In order to prevent the coefficients $a_{j}, j \in \pi_0$ from getting close to zero, the active coefficients were generated via a Bernoulli-Gaussian distribution of the form:
\begin{equation}
a \sim 0.5 \mathcal{N}(1,\sigma^2) + 0.5 \mathcal{N}(-1,\sigma^2),
\label{BerGauss}
\end{equation}

\begin{figure} [h]
\centering
\includegraphics[width=3.0in]{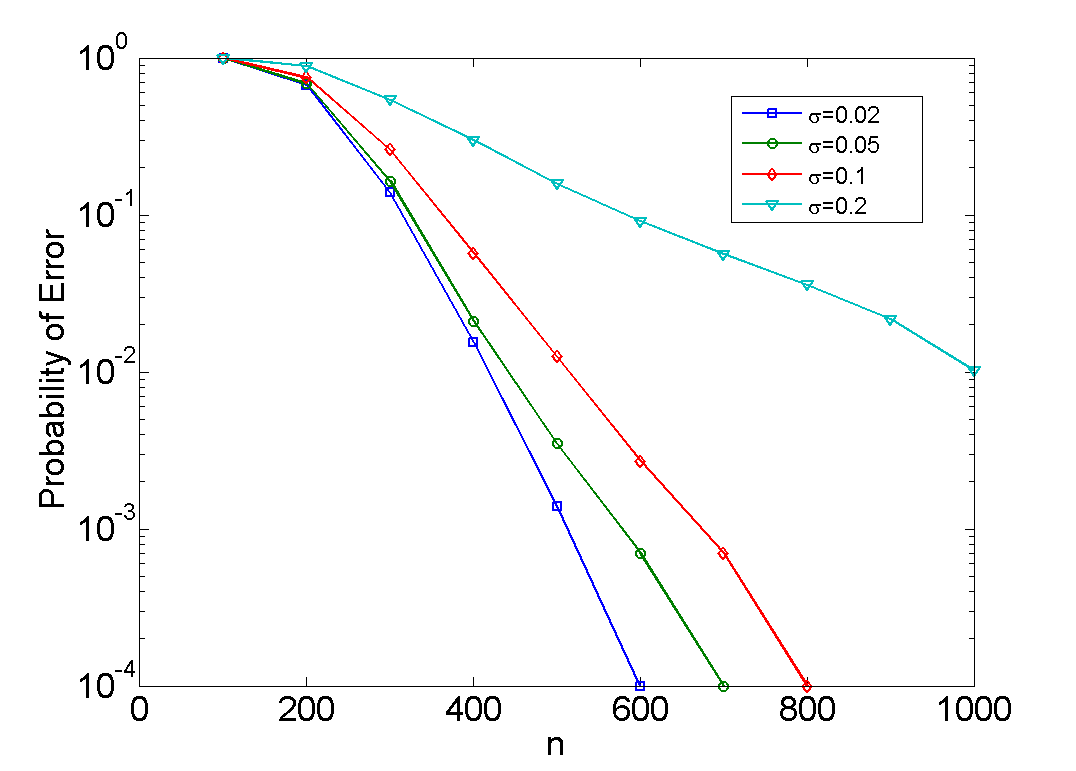}
\caption{\footnotesize Probability of selection error as a function of number of samples for PCS. Probability of selection error is calculated as the ratio of the number of experiments in which the exact support is not recovered over the total number of experiments. The entries of the coefficient matrix are i.i.d. draws from distribution \eqref{BerGauss}. Observe that the probability of selection error decreases at least exponentially with the number of samples. This behavior is consistent with Prop. \ref{Prop:UpperBoundAlg1}.}
\label{fig:UpperBound_logy}
\end{figure}

Figure \ref{fig:UpperBound_logy} shows the estimated probabilities. Each point of the plot is an average of $N=10^4$ experiments. As the value of $\sigma$ decreases the quantity $\rho_{\min}$ defined in \eqref{eq:rhomin} is bounded away from $0$ with high probability and the probability of selection error degrades. As we can see, the FWER decreases at least exponentially with the number of samples. This behavior is consistent with the result in Prop. \ref{Prop:UpperBoundAlg1}.

\textit{d) Application to experimental data.}
We illustrate the proposed SPARCS predictor on the  Predictive Health and Disease data set, which consists of gene expression levels and symptom scores of $38$ different subjects. The data was collected during a challenge study for which  some subjects become symptomatically ill with the H3N2 flu virus \cite{huang2011temporal}. For each subject, the gene expression levels (for $p=12023$ genes) and the clinical symptoms have been recorded at a large number of time points that include pre-inoculation and post-inoculation sample times. Ten different symptom scores were measured. Each symptom score takes an integer value from $0$ to $4$, which measures the severity of that symptom at the corresponding time. 
The goal here is to learn a predictor that can accurately predict the future symptom scores of a subject based on her last measured gene expression levels.

We considered each symptom as a scalar response variable and applied the SPARCS predictor to each symptom separately. In order to do the prediction task, the data used for the SPARCS predictor consists of the samples of the symptom scores for various subjects at $4$ specified time points ($t_1,t_2,t_3,t_4$) and their corresponding gene expression levels measured at the previous time points ($t_1-1,t_2-1,t_3-1,t_4-1$). The number of predictor variables (genes) selected in the first stage is restricted to $100$. Since, the symptom scores take integer values, the second stage uses multinomial logistic regression instead of the OLS predictor. Maximum likelihood estimation is used for computing the multinomial logistic regression coefficients \cite{albert1984existence}. The performance is evaluated by leave-one-out cross validation. To do this, the data from all except one subject are used as training samples and the data from the remaining subject are used as the test samples. The final RMSE is then computed as the average over the $38$ different leave-one-out cross validation trials. In each of the experiments $18$ out of the $37$ subjects of the training set, are used in first stage and all of the $37$ subjects are used in the second stage. It is notable that PCS-SPARCS performs better in predicting the symptom scores for $7$ of the $10$ symptoms whereas SIS-SPARCS and LASSO perform better in predicting the symptom scores for $2$ symptoms and $1$ symptom, respectively.

\begin{table*} \footnotesize
\center
\begin{tabular}{|l|l|l|l|}
  \hline
  Symptom & RMSE: LASSO & RMSE: SIS-SPARCS & RMSE: PCS-SPARCS\\
  \hline
Runny Nose 	&	0.7182	&	0.6896	&	\textbf{0.6559}	\\
Stuffy Nose 	&	0.9242	&	\textbf{0.7787}	&	0.8383	\\
Sneezing 	&	0.7453	&	0.6201	&	\textbf{0.6037}	\\
Sore Throat 	&	0.8235	&	0.7202	&	\textbf{0.5965}	\\
Earache 	&	\textbf{0.2896}	&	0.3226	&	0.3226	\\
Malaise 	&	1.0009	&	\textbf{0.7566}	&	0.9125	\\
Cough 	&	0.5879	&	0.7505	&	\textbf{0.5564}	\\
Shortness of Breath 	&	0.4361	&	0.5206	&	\textbf{0.4022}	\\
Headache 	&	0.7896	&	0.7500	&	\textbf{0.6671}	\\
Myalgia 	&	0.6372	&	0.5539	&	\textbf{0.4610}	\\

  \hline
  Average for all symptoms & 0.6953 & 0.6463 & \textbf{0.6016}\\
  \hline
\end{tabular}

\caption{\footnotesize RMSE of the two-stage LASSO predictor, the SIS-SPARCS predictor and the PCS-SPARCS predictor used for symptom score prediction. The data come from a challenge study experiment that collected gene expression and symptom data from human subjects \cite{huang2011temporal}. Leave-one-out cross validation is used to compute the RMSE values.}
\label{table:PHD}
\end{table*}

\section{Conclusion}

We proposed an online procedure for budget-limited predictor design in high dimensions dubbed two-stage Sampling, Prediction and Adaptive Regression via Correlation Screening (SPARCS). SPARCS is specifically useful in cases where $n \ll p$  and the high cost of assaying all predictor variables justifies a two-stage design: high throughput variable selection followed by predictor construction using fewer selected variables.
We established high dimensional false discovery rates, support recovery guarantees, and optimal stage-wise sample allocation rule associated with the SPARCS online procedure. Simulation and  experimental results showed advantages of SPARCS as compared to LASSO. Our future work includes using SPARCS in a multi-stage framework. We believe that multi-stage SPARCS can further improve the performance of the algorithm while benefiting from high computational efficiency.


\section{Appendix}
\label{sec:appendix}
This section contains three subsections. Section \ref{Appx:Uscores} provides the proof of Lemma \ref{lemma:Phixy}. 
Section \ref{Appx:notation} introduces the necessary notations for the proofs of the remaining propositions. Section \ref{Appx:proofs} gives the proofs for the propositions presented in Sec. \ref{sec:Performance1}.

\subsection{Lemma \ref{lemma:Phixy} and U-score representations}
\label{Appx:Uscores}


Below we present the proof of Lemma \ref{lemma:Phixy} which states that both SIS and PCS methods for discovering the support are equivalent to discovering the non-zero entries of some $p \times 1$ vector $\mathbf \Phi^{xy}$ with representation \eqref{eq:Phixy} by thresholding at a specified threshold.

\noindent{\it Proof of Lemma \ref{lemma:Phixy}:}
Using the U-score representation of the correlation matrices, there exist a $(n-1) \times p$ matrix $\mU^x$ with unit norm columns, and a $(n-1) \times 1$ unit norm vector $\mU^y$ such that \cite{hero2011large,hero2012hub}:
\be
\bR^{xy} = (\mU^x)^T \mU^y.
\label{eq:Rxy2}
\ee
Representation \eqref{eq:Rxy2} immediately shows that SIS is equivalent to discovering non-zero entries of a vector with representation \eqref{eq:Phixy}.
Moreover, we have
\begin{equation}
\bS^{xy}= \bD_{\bS^x}^{\frac{1}{2}} (\mU^x)^T\mU^y (s^y)^{\frac{1}{2}},
\label{eq:Syx}
\end{equation}
and:
\begin{equation}
(\bS^x)^{\dagger}=\bD_{\bS^x}^{-\frac{1}{2}}((\mU^x)^T(\mU^x(\mU^x)^T)^{-2}\mU^x)\bD_{\bS^x}^{-\frac{1}{2}},
\label{eq:Sx}
\end{equation}
where $\bD_{\bA}$ denotes the diagonal matrix obtained by zeroing out the off-diagonals of square matrix $\bA$. We refer the interested reader to \cite{hero2012hub,anderson2003introduction} for more information about the calculations of U-scores.
Using representations \eqref{eq:Syx} and \eqref{eq:Sx}, one can write: \begin{eqnarray}
\hat{Y} &=& ((\bS^x)^{\dagger}\bS^{xy})^T \bX \nonumber \\ &=&
(s^y)^{\frac{1}{2}}(\mU^y)^T(\mU^x(\mU^x)^T)^{-1}\mU^x \bD_{\bS^x}^{-\frac{1}{2}} \bX. \end{eqnarray} Defining
$\tilde{\mU}^x=(\mU^x(\mU^x)^T)^{-1}\mU^x
\bD_{(\mU^x)^T(\mU^x(\mU^x)^T)^{-2}\mU^x}^{-\frac{1}{2}}$, we
have: \begin{eqnarray} \hat{Y}= (s^y)^{\frac{1}{2}} (\mU^y)^T \tilde{\mU}^x
\bD_{(\mU^x)^T(\mU^x(\mU^x)^T)^{-2}\mU^x}^{\frac{1}{2}}  \bD_{\bS^x}^{-\frac{1}{2}} \bX \nonumber \\=
(s^y)^{\frac{1}{2}} (\bH^{xy})^T
\bD_{(\mU^x)^T(\mU^x(\mU^x)^T)^{-2}\mU^x}^{\frac{1}{2}} \bD_{\bS^x}^{-\frac{1}{2}} \bX, \end{eqnarray}
where \be \bH^{xy} =  (\tilde{\mU}^x)^T \mU^y. \label{Hxy} \ee Note
that the columns of the matrix $\tilde{\mU}^x$ lie on $S_{n-2}$ since the  diagonal entries of
the $p \times p$ matrix $(\tilde{\mU}^x)^T \tilde{\mU}^x$ are
equal to one. Therefore, a U-score representation of the generalized OLS solution $\mathbf B^{xy}$ can be obtained as: 
%
 \begin{eqnarray}
\mathbf B^{xy} &=& (\bS^x)^{\dagger} \bS^{xy} \nonumber \\ &=&  \bD_{\bS^x}^{-\frac{1}{2}} \bD_{(\mU^x)^T(\mU^x(\mU^x)^T)^{-2}\mU^x}^{\frac{1}{2}} \bH^{xy}
 (s^y)^{\frac{1}{2}}, \label{eq:CoefMat}\end{eqnarray}
Without loss of generality we can consider the case where $\bD_{\mathbf \Sigma_x} = I_p$. Given the concentration property \eqref{CProperty}, asymptotically we have $\bD_{\bS^x} \rightarrow \bD_{\mathbf \Sigma_x} = I_p$, with probability 1. 
Moreover Assumption \ref{Asmp:coherency} yields the asymptotic relationship $\bD_{(\mU^x)^T(\mU^x(\mU^x)^T)^{-2}\mU^x} = (n-1)^2/p^2 \bI_{p}$. Therefore, finding the largest entries of $\mathbf B^{xy}$ is equivalent to finding the largest entries of $\bH^{xy}$ as the ordering of the entries will asymptotically stay unchanged. This motivates screening for non-zero entries of the vector $\bH^{xy}$ instead of the entries of
$\mathbf B^{xy}$.  
In particular, for a threshold $\rho \in [0,1]$, we can undertake variable selection by discovering the entries of
the vector $\bH^{xy}$ in \eqref{Hxy} that have absolute values at least $\rho$. 
This implies that discovering the support via PCS is equivalent to discovering the non-zero entries of $\bH^{xy}$ in \eqref{Hxy} which admits the representation \eqref{eq:Phixy}. The proof for SIS follows similarly.
 \qed

\subsection{Notations and preliminaries}
\label{Appx:notation}
The following additional notations are necessary for the remaining propositions and the proofs presented in this section.

For arbitrary joint densities
$f_{\bU_i^x,\bU^y}(\bu,\bv), 1 \leq i \leq p$ defined on the Cartesian product
$S_{n-2} \times S_{n-2}$,  define
\be \ol{f_{\bU_{\ast}^x,\bU^y} (\bfu,\bfv)} = \frac{1}{4p} \sum_{i=1}^p \sum_{s,t \in \{0,1\}} f_{\bU_{i}^x,\bU^y}(s \bfu, t \bfv). \ee 
The quantity $\ol{f_{\bU_{\ast}^x,\bU^y} (\bfu,\bfv)}$ is key in determining the expected number of discoveries in screening the entries of the vector $\mathbf \Phi^{xy}$ in \eqref{eq:Phixy}.

In the following propositions, $q$ represents an upper bound on the number of entries
in any row or column of covariance matrix $\mathbf \Sigma_x$ or cross-covariance
vector $\mathbf \Sigma_{xy}$ 
that do not converge to zero as $p \rightarrow \infty$. 
We define $\| \Delta^{xy}_{p,n,q}\|_1$, the
average dependency coefficient, as:
\be
\| \Delta^{xy}_{p,n,q}\|_1 = \frac{1}{p} \sum_{i=1}^{p} \Delta^{xy}_{p,n,q}(i)
\ee
with
\be
&&\Delta_{p,n,q}^{xy}(i) \nonumber = \\ && \left
\|(f_{\bU_{i}^x,\bU^{y} |\bU_{A_q(i)}}-f_{\bU_{i}^x,\bU^{y}})/f_{\bU_{i}^x,\bU^{y}}
\right\|_{\infty}, \ee in which $A_q(i)$ is defined as the
set complement of indices of the $q$-nearest neighbors of $\bU_i^x$ (i.e. the complement of indices of the $q$ entries with largest magnitude in the $i$-th row of $\mathbf \Sigma_x$). Finally, the function $J$ of the joint
density $f_{\bU,\bV}(\bu,\bv)$ is
defined as: \be J(f_{\bU,\bV}) =
|S_{n-2}| \int_{S_{n-2}}
f_{\bU,\bV}(\bw,\bw) d\bw. \label{Jdef} \ee

The function $J(f_{\bU,\bV})$ plays a key role in the asymptotic expression for the mean number of discoveries. 
Note that when observations are independent, by symmetry, the marginal distributions of $U$-scores are exchangeable, i.e.,
\be
f_{\bU}(\bu) = f_{\bU}(\mathbf{\Pi} \bu) ~~~~~~\text{and}~~~~~~ f_{\bV}(\bv) = f_{\bV}(\mathbf{\Pi} \bv),
\ee
for any $(n-1)\times(n-1)$ permutation matrix $\mathbf{\Pi}$. Therefore, the joint distribution $f_{\bU,\bV}$ must yield exchangeable marginals.

We now present two examples for which  $J(f_{\bU,\bV})$ has a closed form expression.

\emph{Example 1.} If the joint distribution $f_{\bU,\bV}$ is uniform over the product $S_{n-2} \times S_{n-2}$,
\be
 J(f_{\bU,\bV}) &=&
|S_{n-2}| \int_{S_{n-2}} \frac{1}{|S_{n-2}|^2} d\bw \nonumber \\ &=& \frac{|S_{n-2}|^2}{|S_{n-2}|^2}=1.
\ee

\emph{Example 2.} Consider the case where the joint distribution $f_{\bU,\bV}$ is separable of the form 
\be
f_{\bU,\bV}(\bu,\bv) = f_{\bU}(\bu)f_{\bV}(\bv),
\ee
i.e., $\bU$ and $\bV$ are independent. Let the marginals be von Mises-Fisher distributions over the sphere $S_{n-2}$
\be
f_{\bU}(\bu) = C_{n-1}(\kappa) \exp(\kappa \boldsymbol \mu^T \bu),~\bu \in S_{n-2},
\ee
in which $\boldsymbol \mu$ and $\kappa \geq 0$ are the location parameter and the concentration parameter, respectively, and $C_{n-1}(\kappa)$ is a normalization constant, calculated as:
\be
C_{n-1}(\kappa) = \frac{\kappa^{(n-1)/2-1}}{(2\pi)^{(n-1)/2}\overline{I}_{(n-1)/2-1}(\kappa)},
\label{eq:Cn}
\ee
where $\overline{I}_m$ is the modified Bessel function of the first kind of order $m$. $\overline{I}_m(x)$ can be computed up to the desired precision using the expansion:
\be
\overline{I}_m(x) = \sum_{l=0}^{\infty} \frac{(x/2)^{2l+n}}{l!\Gamma(l+m+1)},
\label{eq:gammafunc}
\ee
in which $\Gamma(.)$ is the gamma function.

Due to exchangeability of $f_{\bU}(\bu)$, the only two feasible choices for $\boldsymbol \mu$ are $\boldsymbol \mu = \mathbf{1}$ and $\boldsymbol \mu = -\mathbf{1}$, where $\mathbf{1}=[1,1,\ldots,1]^T$. Hence the joint distribution can be written as:
\begin{eqnarray}
&&f_{\bU,\bV}(\bu,\bv) = f_{\bU}(\bu)f_{\bV}(\bv) \nonumber \\ &=& C_{n-1}(\kappa_1) \exp(\kappa_1 \boldsymbol \mu_1^T \bu) C_{n-1}(\kappa_2) \exp(\kappa_2 \boldsymbol \mu_2^T \bv) \nonumber \\ &=& C_{n-1}(\kappa_1) C_{n-1}(\kappa_2) \exp(\kappa_1 \boldsymbol \mu_1^T \bu + \kappa_2 \boldsymbol \mu_2^T \bv) 
\end{eqnarray}
Assuming $\boldsymbol \mu_1 = \alpha_1 \mathbf{1}$ and $\boldsymbol \mu_2 = \alpha_2 \mathbf{1}$, where $\alpha_1,\alpha_2 \in \{-1,1\}$, we obtain:
\begin{eqnarray}
&&f_{\bU,\bV}(\bu,\bv)   \\ \nonumber &=& C_{n-1}(\kappa_1) C_{n-1}(\kappa_2) \exp\left(\mathbf{1}^T (\alpha_1 \kappa_1 \bu + \alpha_2 \kappa_2 \bv) \right).
\end{eqnarray}
This yields:
\be
&& J(f_{\bU,\bV}) \nonumber \\ &=&
|S_{n-2}| \int_{S_{n-2}}
C_{n-1}(\kappa_1) C_{n-1}(\kappa_2) \nonumber \\ && \exp\left( (\alpha_1 \kappa_1 + \alpha_2 \kappa_2) \mathbf{1}^T \bw \right) d\bw \nonumber \\ &=& |S_{n-2}| 
C_{n-1}(\kappa_1) C_{n-1}(\kappa_2) \int_{S_{n-2}} \nonumber \\ && \exp\left( (\alpha_1 \kappa_1 + \alpha_2 \kappa_2) \mathbf{1}^T \bw \right) d\bw \nonumber \\ &=& \frac{|S_{n-2}| C_{n-1}(\kappa_1) C_{n-1}(\kappa_2)}{C_{n-1}(|\alpha_1 \kappa_1 + \alpha_2 \kappa_2|)}. 
\ee 
Therefore, using \eqref{eq:Cn} and \eqref{eq:gammafunc}, $J(f_{\bU,\bV})$ can be computed up to the desired precision.

Further properties as well as intuitive interpretations of $J(f_{\bU,\bV})$ have also been considered in \cite{hero2011large}.

\subsection{Proofs of propositions}
\label{Appx:proofs}

We first prove the following more general version of Prop. \ref{Prop2}. This  generalization can be useful in obtaining an approximate false discovery rates for SPARCS screening stage in cases where the underlying distribution of data is known.

\begin{propositions}
Consider the linear model \eqref{eq:LinearModel} for which Assumption \ref{Asmp:elliptical} is satisfied. Let $\mathbb{U}^x=[\bU_1^x, \bU_2^x,...,\bU_p^x]$ and $\mathbb{U}^y=
[\bU^y]$ be $(n-1)\times p$ and $(n-1) \times 1$ random matrices with unit norm columns. 
Let $\{\rho_p\}_p$ be a sequence of threshold values in $[0,1]$ such
that $\rho_p \rightarrow 1$ as $p\rightarrow \infty$ and
$p(1-\rho_p^2)^{(n-2)/2} \rightarrow
e_{n}$. Throughout this proposition $N^{xy}_{\rho}$ denotes the number of entries of the $p \times 1$ vector $\mathbf G^{xy} = (\mathbb{U}^x)^T \mathbb{U}^y$ whose magnitude is at least $\rho$. We have: \be \lim_{p\rightarrow \infty}
\mathbb E[N^{xy}_{\rho_p}] &=& \lim_{p\rightarrow \infty}
\xi_{p,n,\rho_p} J(\overline{f_{\bU_{*}^x,\bU^y}}) \nonumber \\ &=& \zeta_{n}
\lim_{p\rightarrow \infty} J(\overline{f_{\bU_{*}^x,\bU^y}}), \label{prop1first} \ee where $
\xi_{p,n,\rho_p}=p P_0(\rho,n)
\label{xidef}$ and $ \zeta_{n}= e_{n}
\ra_n/(n-2)$.\\
Assume also that $q=o(p)$ and
that the limit of average dependency coefficient satisfies $\lim_{p\rightarrow \infty}
\| \Delta^{xy}_{p,n,q}\|_1=0$. Then:
\be \mathbb{P}(N_{\rho_p}^{xy}>0) \rightarrow 1-\exp(-\Lambda^{xy}),
\label{eq:Poissonconvcross}\ee
with \be
\Lambda^{xy}= \lim_{p \rightarrow \infty} \mathbb E[N_{\rho_p}^{xy}].
\label{eq:Poissonconvcross2}
\ee
\label{Prop1}
\end{propositions}
\noindent{\it Proof of Prop. \ref{Prop1}:}
Let $d_i^x$ denote the
degree of vertex $X_i$ in part $x$ of the graph ${\mathcal
G}_{\rho}({\mathbf G^{xy}})$. We have: \be N_{\rho}^{xy} =
\sum_{i=1}^{p} d_i^x. \ee The following representation for $d_i^x$ holds: \be d_i^x=I(\bU^y \in
A(r,\bU_i^x)), \ee where $A(r,\bU_i^x)$ is the union of two anti-polar
caps in $S_{n-2}$ of radius $\sqrt{2(1-\rho)}$ centered at $\bU_i^x$
and $-\bU_i^x$.
The following inequality will be helpful: \begin{eqnarray}
\label{exprep2} \mathbb E[d_i^x] =
\int_{S_{n-2}} d\bu \int_{A(r,\bu)} d\bv ~ f_{\bU_{i}^x,\bU^y} (\bu,\bv) \end{eqnarray}
\be \label{expbound2} \leq P_0 a_n M_{1|1}^{yx}, \ee where
$M_{1|1}^{yx} = \max_i \|
f_{\bU^y|\bU_i^x} \|_{\infty}$, and $P_0$ is a simplified notation for  $P_0(\rho,n)$. 
Also for $i \neq j$ we have: \be \label{Qbound} \mathbb E[d_i^x d_j^x] \leq P_0^2 a_n^2 M_{2|1}^{xy}, \ee where
$M_{2|1}^{xy}$ is a bound on the conditional joint densities of the form $f_{\bU_i^x,\bU_j^x|\bU^y}$.\\
Application of the mean value theorem to the integral
representation \eqref{exprep2} yields: \be \label{thetadiffbound}
|\mathbb E[d_i^x] - P_0 J(f_{\bU_i^x,\bU^y })| \leq
\tilde{\gamma}^{yx} P_0 r, \ee where
$\tilde{\gamma}^{yx} = 2a_n^{2}\dot{M}^{yx}_{1|1}$
and $\dot{M}^{yx}_{1|1}$ is a bound on the norm of the
gradient: \be \dot{M}^{yx}_{1|1} = \max_i \| \nabla_{\bU^y}
f_{\bU^y|\bU_i^x}(\bu^y|\bu_i^x)\|_\infty. \ee Using
 \eqref{thetadiffbound} and the
relation $r=O\left((1-\rho)^{1/2}\right)$ we conclude: \begin{eqnarray}
|\mathbb E[d_i^x]- P_0 
J(\overline{f_{\bU_i^x,\bU^y }})| \leq O\left(P_0 (1-\rho)^{1/2}\right).
\end{eqnarray} Summing up over $i$ we conclude:
\begin{eqnarray}
|\mathbb E[N_{\rho}^{xy}]-\xi_{p,n,\rho}
J(\overline{f_{\bU_{*}^x,\bU^y}})|
\leq  O\left(pP_0 (1-\rho)^{1/2}\right)\nonumber \\
 = O\left( \eta_{p}^{xy} (1-\rho)^{1/2} \right), \label{MeanBound}
\end{eqnarray} where $\eta_{p}^{xy} = pP_0$.
This concludes \eqref{prop1first}.\\
To prove the second part of the theorem, we use Chen-Stein method
\cite{arratia1990poisson}. Define the index set
$B^{xy}(i) = \mathcal{N}_{q}^{xy}(i)-\{i\}, 1 \leq i \leq p$, where $\mathcal{N}_{q}^{xy}(i)$ is the set of indices of the $q$-nearest neighbors of $\bU_i^x$. Note that
$|B^{xy}(i)| \leq q$. Assume $N^{*xy}_{\rho}$ is a Poisson random variable with
$\mathbb E[N^{*xy}_{\rho}] = \mathbb E[N^{xy}_{\rho}]$. Using
theorem 1 of \cite{arratia1990poisson}, we have: \be 2 ~\text{max}_A
|\mathbb P(N^{xy}_{\rho} \in A ) -
\mathbb P(N^{*xy}_{\rho} \in A)| \nonumber \\ \leq b_1 + b_2 + b_3, \ee
where: \be b_1 = \sum_{i=1}^p \sum_{i \in
B^{xy}(i)}  \mathbb E[d_i^x] \mathbb E[d_j^x], \\
b_2 = \sum_{i=1}^p \sum_{j \in
B^{xy}(i)}  \mathbb E[d_i^xd_j^x], \label{eq:b2} \ee
and \be b_3 = \sum_{i=1}^p E\left[E\left[ d_i^x - \mathbb E[d_i^x] |
d_j^x : j \in A_{q}(i) \right]\right], \ee where $A_{q}(i)=\left(B^{xy}(i)\right)^c-\{i\}$. Using
the bound \eqref{expbound2}, $\mathbb E[d_i^x]$ is of order $O(P_0)$. Therefore: \be
b_1 \leq O(pk P_0^{2}) =
O((\eta_{p}^{xy})^{2} q/p). \ee Since $i \notin B^{xy}(i)$, applying \eqref{Qbound} to each term of the summation \eqref{eq:b2} gives:
\be b_2 \leq O(p q P_0^2) = O((\eta_{p}^{xy})^{2}q/p). \ee Finally, to
bound $b_3$ we have:
\begin{eqnarray}
&& b_3 = \sum_{i=1}^p  \mathbb E\left[ \mathbb E\left[ d_i^x - \mathbb E[d_i^x] | \bU_{A_{q}(i)} \right]\right]
\nonumber \\ &&= \sum_{i=1}^p
\int_{S_{n-2}^{|A_{q}(i)|}} d\bu_{A_{q}(i)}
\int_{S_{n-2}} d\bu_{i}^x
\int_{A(r,\bfu_{i}^x)} d\bfu^y \nonumber \\
\nonumber  &&\frac{f_{\bU_{i}^{x},\bU^y |
\bU_{A_{q}(i)}}(\bu_{i}^{x},\bu^y |
\bu_{A_{q}(i)}) -
f_{\bU_{i}^{x},\bU^y}(\bu_{i}^{x},\bu^y)}{f_{\bU_{i}^{x},\bU^y}(\bu_{i}^{x},\bu^y)} \times
\nonumber \\ && \nonumber  f_{\bU_{i}^{x},\bU^y}(\bu_{i}^{x},\bu^y)f_{\bU_{A_{q}(i)}}(\bfu_{A_{q}(i)})
\\ &&\leq   O(p P_0 \|\Delta_{p,n,q}^{xy}\|_1)=O(\eta_{p}^{xy} \|\Delta_{p,n,q}^{xy}\|_1).
\end{eqnarray}
Therefore using bound \eqref{MeanBound} we obtain:
\begin{eqnarray} \label{Poissonbound}
&& |\mathbb P(N_{\rho}^{xy} > 0)- (1-\text{exp}(-\Lambda^{xy}))| \leq \nonumber \\
 &&|\mathbb P(N_{\rho}^{xy} > 0)-
(1-\text{exp}(-\mathbb E[N_{\rho}^{xy}]))| \nonumber \\ &+& 
|\text{exp}(-\mathbb E[N_{\rho}^{xy}]) -
\text{exp}(-\Lambda^{xy})| \leq \nonumber \\ && b_1 + b_2 + b_3 \nonumber  +
O(|\mathbb E[N_{\rho}^{xy}] -\Lambda^{xy}|) \leq \nonumber \\  && b_1 + b_2 + b_3 +
O\left( \eta_{p}^{xy} (1-\rho)^{1/2} \right).
\end{eqnarray} Combining this  with the bounds on $b_1, b_2$ and $b_3$, completes the proof of
\eqref{eq:Poissonconvcross}. \qed

In order to obtain stronger bounds, we prove the Prop. \ref{Prop2} under the weakly block-sparse assumption \eqref{eq:bsplustoeplitz}. However the proof for the general case where Assumption \ref{Asmp:coherency} is satisfied follow similarly.

\noindent{\it {Proof of Prop. \ref{Prop2}:}} Proof follows directly from Prop. \ref{Prop1} and Lemma \ref{Lemma:UtildeApprox} presented below. \qed

\begin{lemmas}
Assume the hypotheses of Prop. \ref{Prop2}. Assume also that the correlation matrix $\mathbf{\Omega}_x$ is of the weakly block-sparse from \eqref{eq:bsplustoeplitz} with $d_x = o(p)$.
%
We have:
\be \tilde{\mU}^x=\mU^x(1+O(d_x/p)) . \label{tildeZ}\ee
Moreover, the $2$-fold average function
$J(\ol{f_{\bU_{\ast}^x,\bU^y}})$ and the average dependency coefficient
$\|\Delta_{p,n,q}^{xy}\|$ satisfy \be
J(\ol{f_{\bU_{\ast}^x,\bU^y}})=1 +O((k+d_x)/p), \label{eq:pabrep2}
\ee
\be \|\Delta_{p,n,q}^{xy}\|_1 = 0. \label{eq:pabrep3} \ee
Furthermore, \be J(\ol{f_{\tilde{\bU}_{\ast}^x,\bU^y}})=1 +O(\max
\{d_x/p, d_{xy}/p \})
\label{eq:Jbound2} \ee \be
\|\Delta_{p,n,q}^{\tilde{x}y}\|_1 = O(d_x/p). \label{eq:Delta2}
\ee
\label{Lemma:UtildeApprox}
\end{lemmas}

\noindent{\it {Proof of Lemma \ref{Lemma:UtildeApprox} :}} We have: \be
\tilde{\mU}^x=(\mU^x(\mU^x)^T)^{-1}\mU^x
\bD_{(\mU^x)^T(\mU^x(\mU^x)^T)^{-2}\mU^x}^{-\frac{1}{2}}. \ee 
%
By block sparsity of $\mathbf \Omega_{bs}, \mU^x$ can be partitioned
as: \be \mU^x = [\underline{\mU}^x, \overline{\mU}^x], \ee where
$\underline{\mU}^x=[\underline{\bU}^x_1,\cdots,\underline{\bU}^x_{d_x}]$
are the U-scores corresponding to the dependent block of $\mathbf \Omega_{bs}$ and
$\overline{\mU}^x=[\overline{\bU}^x_1,\cdots,\overline{\bU}^x_{p-d_x}]$ are the remaining U-scores.

Using the law of large numbers for a sequence of correlated variables (see, e.g., Example 11.18 in \cite{severini2005elements}) since the off-diagonal entries of $\mathbf \Omega_x$ that are not in the dependent block converge to $0$ as $|i-j|$ grows, we have
\be
\frac{1}{p-d_x} \overline{\mU}^x (\overline{\mU}^x)^T \rightarrow
\mathbb E[\overline{\bU}^x_1 (\overline{\bU}^x_1)^T] = \frac{1}{n-1}
\bI_{n-1}. \label{lln1}\ee
Since the entries of $1/d_x \underline{\mU}^x
(\underline{\mU}^x)^T$ are bounded by one, we have: \be
\frac{1}{p} \underline{\mU}^x (\underline{\mU}^x)^T =
\bO(d_x/p), \label{bounded_mult}\ee where $\bO(u)$ is an $(n-1)\times(n-1)$ matrix whose
entries are $O(u)$. Hence: \be (\mU^x (\mU^x)^T)^{-1} \mU^x =
(\underline{\mU}^x \left(\underline{\mU}^x)^T + \overline{\mU}^x
(\overline{\mU}^x)^T\right)^{-1} \mU^x \nonumber \\ =
\frac{n-1}{p}(\bI_{n-1}+\bO(d_x/p))^{-1} \mU^x \nonumber \\ =
\frac{n-1}{p} \mU^x (1+O(d_x/p)). \label{Mulbounded}\ee Hence, as
$p \rightarrow \infty$: \be (\mU^x)^T (\mU^x (\mU^x)^T)^{-2} \mU^x = \nonumber \\
= (\frac{n-1}{p})^2 (\mU^x)^T \mU^x (1 + O(d_x/p)). \ee Thus: \be
\bD_{(\mU^x)^T (\mU^x (\mU^x)^T)^{-2} \mU^x}  = \nonumber \\ = \left(\frac{p}{n-1}
\bI_{n-1}(1+O(d_x/p))\right). \label{Dbounded} \ee Combining
\eqref{Dbounded} and \eqref{Mulbounded} concludes \eqref{tildeZ}.

Now we prove relations \eqref{eq:pabrep2}-\eqref{eq:Delta2}.
Define the partition $\{1,\ldots,p\} = {\mathcal D} \cup {\mathcal
D}^c$ of the index set $\{1,\ldots,p\}$,
where ${\mathcal D} = \{i: $
$\bU_i^x$ is asymptotically uncorrelated of $\mU^y \}$.
We have: \be  && J(\ol{f_{\bU_{\ast}^{x}, \bU^y}})= \nonumber \\ && = \frac{1}{4p} \sum_{s,t \in \{-1,1\}}
\label{eq:Crossavgfubi} 
  (\sum_{ i \in {\mathcal D}} + \sum_{ i
\in {\mathcal D}^c}) J(f_{s\bU_{i}^x,t\bU^y}),  \ee and \be
\|\Delta^{xy}_{p,n,q}\|_1 = \frac{1}{p} (\sum_{
i \in {\mathcal D}} + \sum_{ i \in {\mathcal D}^c})
\Delta_{p,n,q}^{xy}(i).\ee But, $J(f_{s\bU_{i}^x,t\bU^y})= 1$ for $i \in  {\mathcal D}$ and
$\Delta_{p,n,q}^{xy}(i) = 0$ for $1 \leq i \leq p$. Moreover, we have $|{\mathcal
D^c}| \leq d_{xy}$, where $d_{xy} = k + d_x$. Therefore,: \be
J(\ol{f_{\bU_{\ast}^{x}, \bU^y}}) = 1 +O(d_{xy}/p). \ee  Moreover, since
$\tilde{\mU}^x = \mU^x \left(1 + O(d_x/p)\right)$, $f_{\tilde{\bU}^x_{i},
\bU^y}=f_{\bU^x_{i},
\bU^y}\left(1+O(d_x/p)\right)$. This
concludes: \be J(\ol{f_{\tilde{\bU}_{\ast}^{x}, \bU^y}}) = 1 +O(\max
\{d_x/p,d_{xy}/p \}), \ee and \be \|\Delta^{\tilde{x}y}_{p,n,q}\|_1 = O(d_x/p). \ee \qed

\noindent{\it Proof of Lemma \ref{Lemma:BlockSparse}:}
By block sparsity of $\mathbf \Omega_{bs}, \mU^x$ can be partitioned
as: \be \mU^x = [\underline{\mU}^x, \overline{\mU}^x], \ee where
$\underline{\mU}^x=[\underline{\bU}^x_1,\cdots,\underline{\bU}^x_{d_x}]$
are the U-scores corresponding to the dependent block of $\mathbf \Omega_{bs}$ and
$\overline{\mU}^x=[\overline{\bU}^x_1,\cdots,\overline{\bU}^x_{p-d_x}]$ are the remaining U-scores.
%
Using relations \eqref{lln1} and \eqref{bounded_mult} we have:
\be \frac{n-1}{p}\mU^x (\mU^x)^T = 
\frac{n-1}{p} \left( \underline{\mU}^x (\underline{\mU}^x)^T + \overline{\mU}^x
(\overline{\mU}^x)^T \right) \nonumber \\ =
\bI_{n-1}+(n-1)\bO(d_x/p). \ee
Noting that $d_x=o(p)$ the result follows.\qed

The following lemma will be useful in the proof of proof of Prop. \ref{Prop:UpperBoundAlg2}.

\begin{lemmas}
\label{Lemma:GeneralCaseLemma} Assume $Z_1, Z_2$ and $Z$ are jointly elliptically contoured distributed random variables from which $n$ joint observations are available. Further assume that the $n \times 3$ matrix $\mathbb Z$ of these observations has an elliptically contoured distribution of the form given in Assumption \ref{Asmp:elliptical}. Let $\rho_1 = \text{Cor}(Z,Z_1)$ and $\rho_2 = \text{Cor}(Z,Z_2)$. Also let $r_1 = \text{SampCor}(Z,Z_1)$ and $r_2 = \text{SampCor}(Z,Z_2)$, be the corresponding sample correlation coefficients. Assume that $|\rho_1|>|\rho_2|$. Then, there exists $C > 0$ and $N$ such that:
\be
\mathbb{P}\left\{ |r_2| > |r_1| \right\} \leq \exp(-Cn),
\ee
for all $n>N$.
\end{lemmas}

We use the following lemma to prove Lemma \ref{Lemma:GeneralCaseLemma}.
\begin{lemmas}
Let $\bU$ and $\bV$ be two independent uniformly distributed random vectors on $S_{n-2}$. For any fixed $\epsilon > 0$, there exists $C>0$ such that:
\be
\mathbb{P}\{|\bU^T\bV| > \epsilon \} \leq \exp(-Cn). 
\ee  
\label{Lemma:SphereCap}
\end{lemmas}
\noindent{\it Proof of Lemma \ref{Lemma:SphereCap}:}
Without loss of generality assume $U=[1,0,\ldots,0]^T$. We have 
\be
\{|\bU_2^T\bU_1| > \epsilon \} = \{ |v_1| > \epsilon \},
\ee
in which $v_1$ is the first entry of the vector $\bV$. Using the formula for the area of spherical cap \cite{li2011concise} we obtain
\be
\mathbb{P}\{|\bU_2^T\bU_1| > \epsilon \} =  I_{\lambda}(n/2,1/2),
\ee
where $\lambda = 1-\epsilon^2$, and \begin{equation}
I_x(a,b)=\frac{\int_{0}^{x}t^{a-1}(1-t)^{b-1}dt}{\int_{0}^{1}t^{a-1}(1-t)^{b-1}dt}
\end{equation}
is the regularized incomplete beta function. Note that:
\be
&& 1/I_{\lambda}(n/2,1/2) = \nonumber \\&& = \frac{\int_{0}^{\lambda}t^{(n-2)/2}/\sqrt{1-t}dt + \int_{\lambda}^{1}t^{(n-2)/2}/\sqrt{1-t}dt}{\int_{0}^{\lambda}t^{(n-2)/2}/\sqrt{1-t}dt}  \nonumber \\ &&= 1+\frac{\int_{\lambda}^{1}t^{(n-2)/2}/\sqrt{1-t}dt}{\int_{0}^{\lambda}t^{(n-2)/2}/\sqrt{1-t}dt} \nonumber \\ &&\geq 1+\frac{\int_{\lambda}^{1}t^{(n-2)/2}/\sqrt{1-\lambda}dt}{\int_{0}^{\lambda}t^{(n-2)/2}/\sqrt{1-\lambda}dt}  \nonumber \\ && = 1 + \frac{1-\lambda^{n/2}}{\lambda^{n/2}} = (\sqrt{\lambda})^n.
\ee
Therefore by letting $C = -\frac{1}{2}\log(\lambda) = -\frac{1}{2}\log(1-\epsilon^2)$ we obtain
\be
\mathbb{P}\{|\bU_2^T\bU_1| > \epsilon \} \leq \exp(-Cn).
\ee
\qed


\noindent{\it Proof of Lemma \ref{Lemma:GeneralCaseLemma}:}
Let $\bZ = [Z_2, Z_1, Z]^T$. Assume $\bZ$ follows an elliptically contoured density function of the form $f_{\bZ}(\bz)= |\mathbf \Sigma_z|^{-1/2}
g\left((\bz-\boldsymbol \mu_z)^T {\mathbf \Sigma_z}^{-1} (\bz-\boldsymbol \mu_z)\right)$. Without loss of generality assume $\Var(Z_1)=\Var(Z_2)=\Var(Z)=1$.
Using a Cholesky factorization we can represent $Z_1,Z_2$ and $Z$ as linear combination of uncorrelated random variables $W_1,W_2$ and $W$ which follow a spherically contoured distribution:

\be
\begin{bmatrix}
        Z_2           \\[0.3em]
        Z_1 \\[0.3em]
        Z 
      \end{bmatrix} = 
\begin{bmatrix}
        1 & 0 & 0           \\[0.3em]
        a & b           & 0 \\[0.3em]
        c           & d & e
      \end{bmatrix} \times \begin{bmatrix}
        W_2           \\[0.3em]
        W_1 \\[0.3em]
        W 
      \end{bmatrix}
      \label{eq:matrixmult}
\ee
where \be \rho_1 = ac + bd, \label{Chol1} \\ \rho_2 = c, \label{Chol2} \\ a^2+b^2 = 1, \label{Chol3} \ee and \be c^2+d^2+e^2 = 1. \label{Chol4} \ee
Let $\bW=[W_2,W_1,W]^T$. 
Since $\bW$ follows a spherically contoured distribution, it has a stochastic representation of the form $\bW = R \bU$, where $R$ has a marginal density $f_{R}(r) = \alpha h(r^2)r^2$, in which $\alpha$ is a normalizing constant. Moreover $\bU$ is independent of $R$ and the distribution of $\bU$ does not depend on the function $h$ (see, e.g., Chapter 2 in \cite{anderson2003introduction} for more details about such stochastic representation). 
Now let $\bU^z_1,\bU^z_2$ and $\bU^z$ denote the U-scores corresponding to $n$ independent samples of $Z_1,Z_2$ and $Z$, respectively. Then under Assumption \ref{Asmp:elliptical}, as these U-scores are invariant to translation and scale on the $n$ samples of $Z_1, Z_2, Z$, the joint distribution of the U-scores does not depend on $g$ and without loss of generality the $n$ samples can be assumed to be i.i.d. Gaussian \cite{anderson1992nonnormal}. Similarly, let $\bU^w_1,\bU^w_2$ and $\bU^w$ denote the U-scores corresponding to $W_1,W_2$ and $W$, respectively. Using \eqref{eq:matrixmult} we have the following relations:
\be
\bU_2^z &=& \bU^w_2, \nonumber \\
\bU_1^z &=& (a\bU^w_2 + b\bU^w_1) / \|  a\bU^w_2 + b\bU^w_1  \|_2, \nonumber \\
\bU^z &=& (c\bU^w_2 + d\bU^w_1 + e\bU^w) / \nonumber \\ && ~~~~~~\|  c\bU^w_2 + d\bU^w_1 + e\bU^w \|_2. 
\ee
Hence
\be
&& r_1 = (\bU^z)^T \bU_1^z  = \nonumber \\ && \frac{1}{ \|  c\bU^w_2 + d\bU^w_1 + e\bU^w \|_2 \|  a\bU^w_2 + b\bU^w_1  \|_2} \times \nonumber \\ && \Big( ac + bd + bc (\bU_2^w)^T\bU^w_1 + ad (\bU_1^w)^T\bU^w_2 \nonumber \\ && + ae (\bU^w)^T\bU^w_2 + be (\bU^w)^T\bU^w_1 \Big),
\ee
and
\be
r_2 &=& (\bU^z)^T \bU^z_2 \nonumber \\ &=& \frac{c + d (\bU_1^w)^T\bU^w_2 + e (\bU^w)^T\bU^w_2} { \|  c\bU^w_2 + d\bU^w_1 + e\bU^w \|_2 }.
\ee
Now let $E = \{|r_2|>|r_1|\} $. We have:
\be
&&E = \big\{ |\bU^T\bU_2|>|\bU^T\bU_1| \big\} = \nonumber \\ && \Big\{ \|  a\bU^w_2 + b\bU^w_1  \|_2 \Big|c + d (\bU_1^w)^T\bU^w_2 + e  (\bU^w)^T\bU^w_2\Big| \nonumber \\ && > \Big| ac + bd + bc (\bU_2^w)^T\bU^w_1 + ad (\bU_1^w)^T\bU^w_2 + \nonumber \\ && + ae (\bU^w)^T\bU^w_2 + be (\bU^w)^T\bU^w_1 \Big| \Big\}. 
\ee
Since  \be  \|  a\bU^w_2 + b\bU^w_1  \|_2 &=& \sqrt{(a\bU^w_2 + b\bU^w_1)^T(a\bU^w_2 + b\bU^w_1)} \nonumber \\ =&&  \sqrt{a^2+b^2 + 2ab(\bU_2^w)^T\bU^w_1}  \nonumber \\ =&& \sqrt{1 + 2ab(\bU_2^w)^T\bU^w_1} \nonumber \\ \leq && 1 + 2|ab|.|(\bU_2^w)^T\bU^w_1|, \ee 
and, by using triangle inequality, we have
\be
&&E \subseteq \Big\{2|abc|.|(\bU_2^w)^T\bU^w_1|^2 + \nonumber \\ && 2|e|.|(\bU^w)^T\bU^w_2|.|(\bU_2^w)^T\bU^w_1| + \nonumber \\ && |ad+bc|.|(\bU_2^w)^T\bU^w_1| +  |ae|.|(\bU^w)^T\bU^w_1|  + \nonumber \\ && |be|.|(\bU^w)^T\bU^w_1| > |ac+bd|-|c| \Big \} \nonumber \\ &&\subseteq  \big \{2|abc|.|(\bU_2^w)^T\bU^w_1|^2 > |ac+bd|-|c| \big \} \bigcup \nonumber \\ && \big \{ 2|e|.|(\bU^w)^T\bU^w_2|.|(\bU_2^w)^T\bU^w_1| > |ac+bd|-|c| \big \}  \bigcup \nonumber \\ &&\big\{ |ad+bc|.|(\bU_2^w)^T\bU^w_1| > |ac+bd|-|c| \big\} \bigcup \nonumber \\ &&\big\{ |ae|.|(\bU^w)^T\bU^w_1| > |ac+bd|-|c| \big\} \bigcup \nonumber \\ && \big\{ |be|.|(\bU^w)^T\bU^w_1| > |ac+bd|-|c| \big\}  \nonumber \\  &&\subseteq \big\{|(\bU_2^w)^T\bU^w_1| > (|ac+bd|-|c|)/2|abc|\big\} \bigcup \nonumber \\ &&\big\{ |(\bU_2^w)^T\bU^w_1| > (|ac+bd|-|c|)/2|e| \big\} \bigcup \nonumber \\ && \big\{ |(\bU_2^w)^T\bU^w_1| > (|ac+bd|-|c|)/|ad+bc| \big\} \bigcup \nonumber \\ && \big\{ |(\bU^w)^T\bU^w_1| > (|ac+bd|-|c|)/|ae| \big\} \bigcup \nonumber \\ && \big\{ |(\bU^w)^T\bU^w_1| > (|ac+bd|-|c|)/|be| \big\}.
\ee
Note that by assumption $|ac+bd| = |\rho_1| > |\rho_2| =  |c|$. Now by Lemma \ref{Lemma:SphereCap} we get
\be
\mathbb{P}(E) \leq 5 \exp(-\alpha n),
\ee
with
\be
\alpha &=& \frac{|ac+bd|-|c|}{\max\left\{ 2|abc|,2|e|,|ad+bc|,|ae|,|be|   \right\}} \nonumber \\ &\geq& \frac{\rho_1-\rho_2}{2},
\ee
where the last inequality is obtained via equations \eqref{Chol1}-\eqref{Chol4}. Letting $C = (\rho_1-\rho_2)/3$ and $N=12/(\rho_1-\rho_2)$ we have
\be
\mathbb{P}(E) = \mathbb{P}\{|r_2|>|r_1|\} \leq \exp(-Cn),\ee for $n> N$. \qed 

\noindent{\it Proof of Proposition \ref{Prop:UpperBoundAlg2}:}
Since $\mathbb{P}\left( \pi_0 \subseteq S \right)$ increases as the size of the recovered set $S$ increases, it suffices to prove the proposition for $l = k$. Define an auxiliary random variable $X_{\text{ax}}$ such that $\text{Cor}(Y,X_{\text{ax}})= \left(\max_{j \in \{1,\cdots,p \} \backslash \pi_0}|\rho_{yj}| + \min_{i \in \pi_0}|\rho_{yi}|\right)/2$. Note that by Assumption \ref{Asmp:rhomin} $\max_{j \in \{1,\cdots,p \} \backslash \pi_0}|\rho_{yj}| < \text{Cor}(Y,X_{\text{ax}}) < \min_{i \in \pi_0}|\rho_{yi}|$. For $l = k$ we have:
\be
&&\mathbb{P}\left( \pi_0 \nsubseteq S \right) = \mathbb{P}\left( \pi_0 \neq S \right) \nonumber \\ &&\leq \mathbb{P} \Bigg( \bigcup_{i \in \pi_0} \{|r_{yi}|<|\text{SampCor}(Y,X_{\text{ax}})| \} \nonumber \\ && \bigcup_{j \in \{1,\ldots, p\} \backslash \pi_0} \{|r_{yj}| > |\text{SampCor}(Y,X_{\text{ax}})| \} \Bigg)~~~ \\  &&\leq \sum_{i \in \pi_0} \mathbb{P} \Big( |r_{yi}|<|\text{SampCor}(Y,X_{\text{ax}})| \Big) + \nonumber \\ \nonumber && \sum_{j \in \{1,\ldots, p\} \backslash \pi_0} \mathbb{P} \Big( |r_{yj}|>|\text{SampCor}(Y,X_{\text{ax}})| \Big).
\label{ineq:exponentialbound}
\ee
Now since Assumptions \ref{Asmp:elliptical} and \ref{Asmp:rhomin} are satisfied, by Lemma \ref{Lemma:GeneralCaseLemma} there exist constants $C_i >0, 1 \leq i \leq p$ and a constant $N$ such that
\be
&&\mathbb{P}\left( \pi_0 \neq S \right) \nonumber \\ && \leq \sum_{i \in \pi_0} \exp(-C_in) + \sum_{j \in \{1,\ldots, p\} \backslash \pi_0} \exp(-C_jn) \nonumber \\ &&\leq
p \exp(-C_{\min}n),~~~~ \forall n>N,
\ee
in which $C_{\min} = \min_{1 \leq i \leq p} C_i = \rho_{\min}/6$. Hence by letting $C = 2/C_{\min} = 12/\rho_{\min}$ and $n = C \log p$ we have:
\be
\mathbb{P}\left( \pi_0 \neq S \right) \leq \frac{1}{p},
\ee
and
\be
\mathbb{P}\left( \pi_0 = S \right)  = 1 - \mathbb{P}\left( \pi_0 \neq S \right) \geq 1 - \frac{1}{p},
\ee
which completes the proof.
\qed

\noindent{\it Proof of Proposition \ref{Prop:UpperBoundAlg1}:}
We only provide a proof sketch here. By Assumption \ref{Asmp:coherency} we have \be \mU^x (\mU^x)^T = \frac{p}{n-1} \left(\bI_{n-1}+ \textbf o(1) \right). \ee 
Therefore:
\be
\left(\mU^x (\mU^x)^T\right)^{-1} = \frac{n-1}{p} \left( \bI_{n-1} + \textbf o(1) \right).
\ee
Since columns of $\mU^x$ have unit norm we obtain:
 \be (\mU^x (\mU^x)^T)^{-1} \mU^x =
\frac{n-1}{p} \mU^x (1+o(1)), \ee and \be && (\mU^x)^T (\mU^x (\mU^x)^T)^{-2} \mU^x = \nonumber \\
&&~~~~~~ (\frac{n-1}{p})^2 (\mU^x)^T \mU^x (1 + o(1)). \ee This yields \be
\bD_{(\mU^x)^T (\mU^x (\mU^x)^T)^{-2} \mU^x} = (\frac{n-1}{p})^2
\bI_{p}(1+o(1)), \ee
which implies \be \tilde{\mU}^x&=&(\mU^x(\mU^x)^T)^{-1}\mU^x
\bD_{(\mU^x)^T(\mU^x(\mU^x)^T)^{-2}\mU^x}^{-\frac{1}{2}} \nonumber \\ &=& \mU^x(1+o(1)). \ee
where, by the concentration assumption, with high probability the term $o(1)$ decays to 0 exponentially fast. Therefore screening the entries of $\bB^{xy}$ or $\bH^{xy}$ is asymptotically equivalent to selecting the support via thresholding the entries of $(\mU^x)^T\mU^y$, i.e., the sample correlation coefficients. Therefore the proof follows from Prop. \ref{Prop:UpperBoundAlg2}.
 \qed

\noindent{\it Proof of Proposition \ref{Prop:UpperBound2}:} First we consider a two-stage predictor similar to the one introduced in Sec. \ref{sec:SPARCS} with the difference that the $n$ samples which are used in stage $1$ are not used in stage $2$. Therefore, there are $n$ and $t-n$ samples used in the first and the second stages, respectively. We represent this two-stage predictor by $n|(t-n)$. Similarly, $n|t$ denotes the SPARCS algorithm which uses $n$ samples at the first stage and all of the $t$ samples at the second stage. The asymptotic results for the $n|(t-n)$ two-stage predictor will be shown to hold as well for the $n|t$ two-stage predictor.

Using inequalities of the form \eqref{ineq:exponentialbound} and the union bound, it is straightforward to see that for any subset $\pi \neq \pi_0$ of $k$ elements of $\{1,\cdots,p \}$, the probability that $\pi$ is the outcome of variable selection via SPARCS, is
bounded above by $p c_{\pi}^n$, in which $0 < c_{\pi}<1$ is a constant that is bounded above by $\exp(-C_{\min})$. 
The expected MSE of the $n|(t-n)$ algorithm can be written as:
\begin{equation}
\mathbb E[\text{MSE}] = \sum_{\pi \in S_k^p, \pi \neq \pi_0} \mathbb P(\pi)\mathbb E[\text{MSE}_{\pi}] +
\mathbb P(\pi_0)\mathbb E[\text{MSE}_{\pi_0}],
\end{equation}
where $S_k^p$ is the set of all $k$-subsets of $\{1,\cdots,p\}$, $\mathbb P(\pi)$ is the probability that the outcome of variable
selection via SPARCS is the subset $\pi$, and $\text{MSE}_{\pi}$ is the MSE of OLS  stage when the indices of the selected variables are the elements of $\pi$.
Therefore the expected MSE is upper bounded as below:
\begin{eqnarray}
\mathbb E[\text{MSE}] \leq  (1-p c_0^n)\mathbb E[\text{MSE}_{\pi}] \nonumber + \\  + p \sum_{\pi \in S_k^p, \pi \neq \pi_0}
 c_{\pi}^n \mathbb E[\text{MSE}_{\pi}],
\end{eqnarray}
where $c_0$ is a constant which is upper bounded by $\exp(-C_{\min})$. 
It can be shown that if there is at least one wrong variable
selected ($\pi \neq \pi_0$), the OLS estimator is biased and the expected MSE
converges to a positive constant $M_{\pi}$ as $(t-n) \rightarrow
\infty$. When all the variables are selected correctly (subset
$\pi_0$), MSE goes to zero with rate $O(1/(t-n))$. Hence:
\be
&&\mathbb E[\text{MSE}] \leq \nonumber \\&& (1-p c_0^n) O(1/(t-n)) + p \sum_{\pi \in S_k^p, \pi \neq \pi_0} c_{\pi}^n
M_{\pi}  \nonumber \leq \\ && (1-p c_0^n) C_2/(t-n) + p^{k+1} C_1 C^n,~~~~~~~~~~~~~~~~~ \label{upper_objective}
\ee
where $C,C_1$ and $C_2$ are constants that do not depend on $n$ or $p$
but depend on the quantities $\sum_{j \in \pi_0} a_{j}^2$ and $\min_{j \in \pi_0} |a_{j}| /\sum_{l \in \pi_0} |a_{l}|$. Note that $C = \max_{\pi \in S_k^p, \pi \neq \pi_0} c_{\pi} \leq \exp(-C_{\min})$. This quantity is an increasing function $\rho_{\min}$. 

On the other hand since at most $t$ variables could be used in OLS
stage, the expected MSE is lower bounded:
\begin{equation}
\mathbb E[\text{MSE}] \geq \Theta (1/t).
\label{eq:lowerb}
\end{equation}



It can be seen that the minimum of \eqref{upper_objective} as a
function of $n$, subject to the constraint \eqref{eq:assaycostcond}, happens for
$n=O(\log t)$ if $ c \log t \leq \frac{\mu - tk}{p-k}$ with $c = -1/\log C$ (therefore, similar to $C$, $c$ is increasing in $\rho_{\min}$);
otherwise it happens for 0. 
If $ \Theta(\log t) \leq \frac{\mu - tk}{p-k}$, the minimum value
attained by the upper bound \eqref{upper_objective} is $\Theta(1/t)$ which is
as low as the lower bound \eqref{eq:lowerb}. This shows that for large $t$, the optimal number of samples that should be assigned to the SPARCS stage of the $n|(t-n)$ predictor is $n=O(\log t)$. As $t
\rightarrow \infty$, since $n=O(\log t)$, the MSE of the $n|t$ predictor proposed in
Sec. \ref{sec:SPARCS} converges to the MSE of the $n|(t-n)$ predictor.
Therefore, as $t \rightarrow \infty$, $n=O(\log t)$ becomes optimal for the $n|t$ predictor as well. \qed 

\bibliography{Refs}
\bibliographystyle{sieee}

\newpage

\begin{IEEEbiography}[{\includegraphics[width=1in,height=1.25in,clip,keepaspectratio]{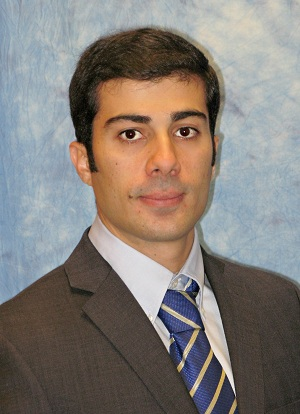}}]{Hamed Firouzi}
received the B.Sc. degrees in Electrical Engineering and Pure Mathematics from Sharif University of Technology, Tehran, Iran, in 2009, and the M.Sc. degrees in Financial Engineering and Applied  Mathematics, from the University of Michigan, Ann Arbor, MI, USA, in 2014. He received the PhD degree in Electrical Engineering from the University of Michigan, Ann Arbor, MI, USA, in 2015. He is currently a  quantitative modeler at the Goldman Sachs Group, Inc. His research interests include predictive modeling, machine learning, pattern recognition, data science, statistical signal processing, and financial modeling.
\end{IEEEbiography}

\begin{IEEEbiography}
[{\includegraphics[width=1in,height=1.25in,clip,keepaspectratio]{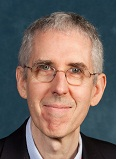}}]{Alfred O. Hero III}
is the John H. Holland Distinguished University Professor of Electrical Engineering and Computer Science and the R. Jamison and Betty Williams Professor of Engineering at the University of Michigan, Ann Arbor. He is also the Co-Director of the University's Michigan Institute for Data Science (MIDAS). His primary appointment is in the Department of Electrical Engineering and Computer Science and he also has appointments, by courtesy, in the Department of Biomedical Engineering and the Department of Statistics. He received the B.S. (summa cum laude) from Boston University (1980) and the Ph.D from Princeton University (1984), both in Electrical Engineering. He is a Fellow of the Institute of Electrical and Electronics Engineers (IEEE). He has served as President of the IEEE Signal Processing Society and as a member of the IEEE Board of Directors. He has received numerous awards for his scientific research and service to the profession including the IEEE Signal Processing Society Technical Achievement Award in 2013 and the 2015 Society Award, which is the highest career award bestowed by the IEEE Signal Processing Society. Alfred Hero's recent research interests are in high dimensional spatio-temporal data, multi-modal data integration, statistical signal processing, and machine learning. Of particular interest are applications to social networks, network security and forensics, computer vision, and personalized health.
\end{IEEEbiography}

\begin{IEEEbiography}[{\includegraphics[width=1in,height=1.25in,clip,keepaspectratio]{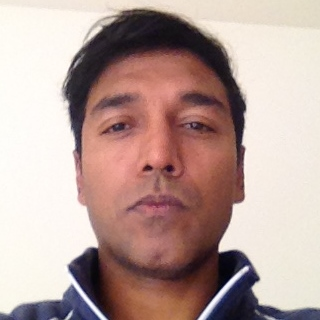}}]{Bala Rajaratnam}
received the B.Sc.(Hons)/M.Sc. degrees from the University of the Witwatersrand, Johannesburg, South Africa, in 1999 and the M.S./Ph.D. degrees from Cornell University, Ithaca, NY, USA, in 2006. He is a faculty member at the Department of Statistics at  Stanford University, Stanford and the University of California Davis,  CA, USA. He is also a visiting professor at the University of Sydney, Australia. His research interests include graphical models, machine learning, data science, high-dimensional inference, signal processing, spatio–temporal and environmental modeling, financial engineering, positivity and the mathematics of networks. Mr. Rajaratnam is the recipient of several awards and recognitions including two federal CAREER awards, the National Science Foundation (NSF) CAREER Award, and the Defense Advanced Research Projects Agency (DARPA) Young Faculty Award.
\end{IEEEbiography}

\end{document}